\documentclass{article}

\usepackage{arxiv}

\usepackage[T1]{fontenc}
\usepackage[utf8]{inputenc}
\usepackage{authblk}
\usepackage{hyperref}       
\usepackage{url}            
\usepackage{booktabs}       
\usepackage{amsfonts}       
\usepackage{nicefrac}       
\usepackage{microtype}      
\usepackage{lipsum}
\usepackage{fancyhdr}       
\usepackage{graphicx}       

\usepackage{graphicx}%
\usepackage[caption=false,font=normalsize,labelfont=sf,textfont=sf]{subfig}
\usepackage{multirow}%
\usepackage{amsmath,amssymb,amsfonts}%
\usepackage{amsthm}%
\usepackage{mathrsfs}%
\usepackage[title]{appendix}%
\usepackage{xcolor}%
\usepackage{textcomp}%
\usepackage{manyfoot}%
\usepackage{booktabs}%
\usepackage{listings}%
\usepackage{todonotes}

\usepackage{makecell}

\usepackage{colortbl}
\usepackage{xcolor}
\definecolor{tabgray}{gray}{0.90}

\usepackage{algorithm}%
\usepackage{algorithmic}%

\usepackage{comment}
\usepackage{soul}
\usepackage{mathtools}
\usepackage[numbers,sort&compress]{natbib}

\usepackage{ulem}
\usepackage{longtable}
\usepackage{multirow}
\usepackage{array}

\newtheorem{proposition}{Proposition}%

\title{Kernel PCA for Out-of-Distribution Detection:\\Non-Linear Kernel Selection and Approximation}
\author[1,2]{Kun Fang}
\author[3]{Qinghua Tao}
\author[1]{Mingzhen He}
\author[4]{Kexin Lv}
\author[5]{Runze Yang}
\author[2]{Haibo Hu}
\author[1]{Xiaolin Huang}
\author[1]{Jie Yang}
\author[5]{Longbing Cao}

\affil[1]{Department of Automation, Shanghai Jiao Tong University
}

\affil[2]{Department of Electrical and
Electronic Engineering, The Hong Kong Polytechnic University
}

\affil[3]{School of Automation, Beijing Institute of
Technology
}

\affil[4]{China Mobile (Shanghai) Information and
Communication Technology Co., Ltd.
}

\affil[5]{School of Computing, Macquarie University
}

\begin{document}

\maketitle

\begin{abstract}
Out-of-Distribution (OoD) detection is vital for the reliability of deep neural networks, the key of which lies in effectively characterizing the disparities between OoD and  In-Distribution (InD) data.
In this work, such  disparities are exploited through a fresh perspective of \textit{non-linear feature subspaces}.
That is, a discriminative non-linear subspace is learned from InD features to capture representative patterns of InD, while informative patterns of OoD features cannot be well captured in such a subspace due to their different distribution. 
Grounded on this perspective, we exploit the deviations of InD and OoD features in such a non-linear subspace for effective OoD detection.
To be specific, we leverage the framework of Kernel Principal Component Analysis (KPCA) to attain the discriminative non-linear subspace and deploy the reconstruction error on such subspace to distinguish InD and OoD data.
Two challenges emerge: {\textit{(i)}} the learning of an effective non-linear subspace, i.e., the selection of kernel function in KPCA, and {\textit{(ii)}} the computation of the kernel matrix with large-scale InD data.
For the former, we reveal two vital non-linear patterns that closely relate to the InD-OoD disparity, leading to the establishment of a Cosine-Gaussian kernel for constructing the subspace.
For the latter, we introduce two techniques to approximate the Cosine-Gaussian kernel with significantly cheap computations. 
In particular, our approximation is further tailored by incorporating the InD data confidence, which is demonstrated to promote the learning of discriminative subspaces for OoD data.
Our study presents new insights into the non-linear feature subspace for OoD detection and contributes practical explorations on the associated kernel design and efficient computations, yielding a KPCA detection framework with distinctively improved efficacy and efficiency.
\end{abstract}

\section{Introduction}
\label{sec:intro}
With{\let\thefootnote\relax\footnotetext{kun.fang@polyu.edu.hk}} the rapid advancement of the powerful learning abilities of Deep Neural Networks (DNNs) \cite{ho2020denoising,ouyang2022training}, the trustworthiness of DNNs in security-sensitive scenarios has attracted considerable attention in recent years \cite{liu2022trustworthy,barrett2023identifying}.
Generally, samples from the training set and  test set of DNNs are viewed as data from some In Distribution (InD) $\mathbb{P}_{\rm in}$, while samples from other data sets are regarded as those  from a different distribution $\mathbb{P}_{\rm out}$, i.e., Out-of-Distribution (OoD) data.
In practical deployments, DNNs trained on InD data can encounter OoD data and yield unreliable results with potential risks.
Therefore, detecting whether a new sample is from $\mathbb{P}_{\rm in}$ or $\mathbb{P}_{\rm out}$ has been a valuable research topic in trustworthy deep learning, namely OoD detection \cite{yang2024generalized}.

\begin{figure*}[t]
    \centering
    
    \subfloat[T-SNE of $\boldsymbol{z}$ and PCA reconstruction errors.]{
    \label{fig:intro-pca}
    \includegraphics[width=0.48\linewidth]{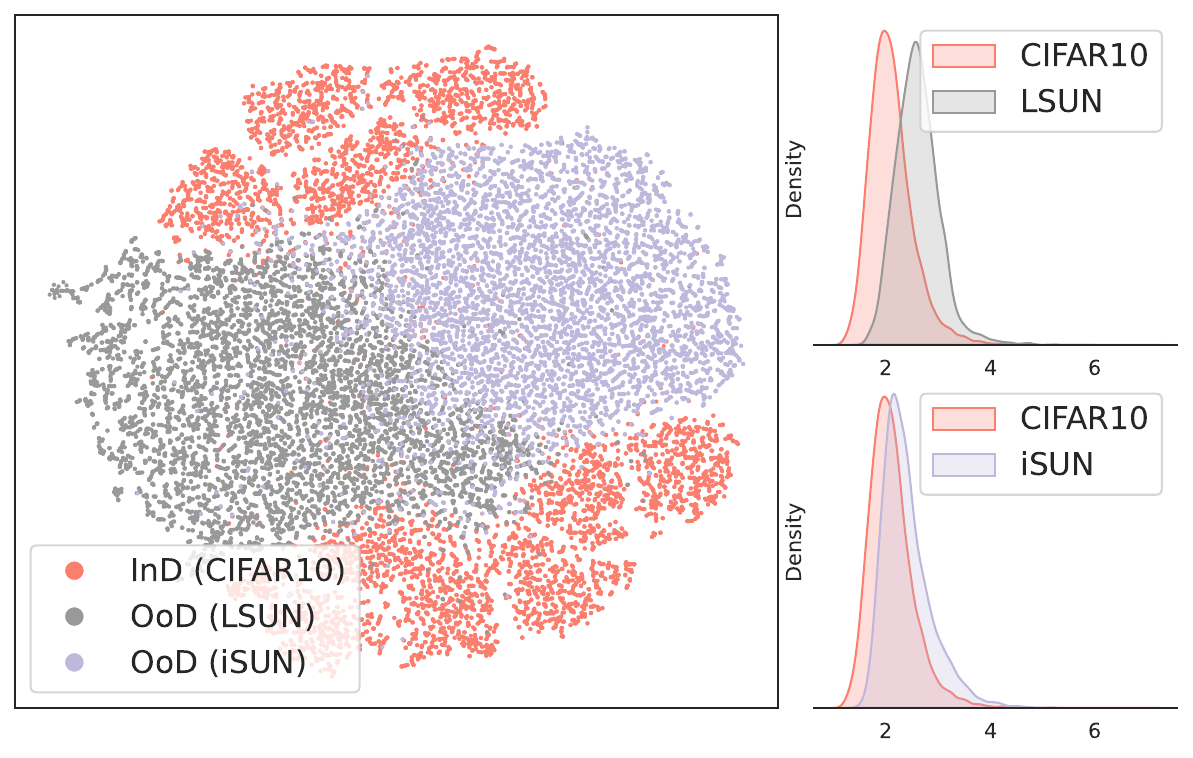}}
    \hfil
    \subfloat[T-SNE of $\Phi(\boldsymbol{z})$ and KPCA reconstruction errors.]{
    \label{fig:intro-kpca}
    \includegraphics[width=0.48\linewidth]{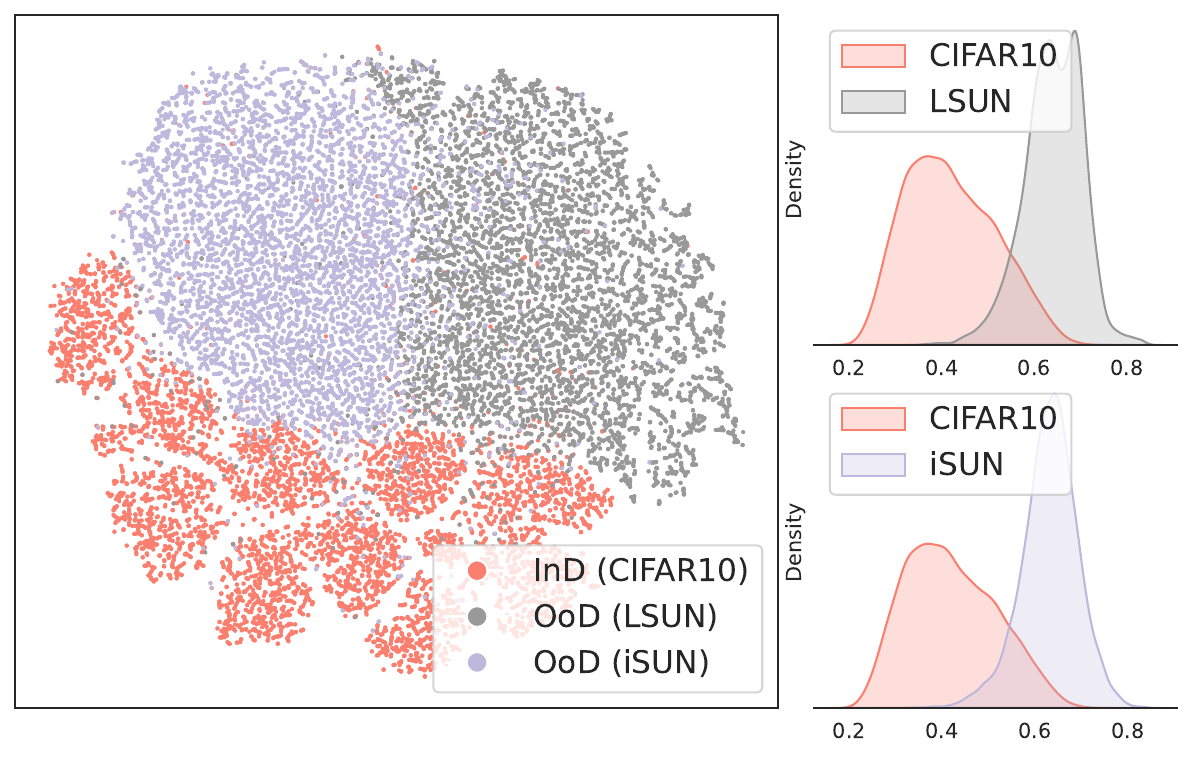}}
    
    \caption{The t-SNE \cite{van2008visualizing} visualization on the original features $\boldsymbol{z}$ (left) and the features $\Phi(\boldsymbol{z})$ in subspace (right). 
    Our KPCA detection method alleviates the linear inseparability between InD and OoD features in the original $\boldsymbol{z}$-space via the mapping $\Phi$ with substantially improved OoD detection performance, illustrated by the distinguishable reconstruction errors.}
    \label{fig:intro}
\end{figure*}

Great efforts have been taken to put forward the research on OoD detection from varied perspectives. 
The key idea of those works lies in the proper use of different responses from DNNs, i.e., the logits \cite{liu2020energy,wang2023wood}, gradients \cite{huang2021importance,wu2024low}, and features \cite{sun2021react,sun2022out,guan2023revisit}, to characterize the disparity between InD and OoD data.
Among those utilized responses, predictive logits merely encode limited and coarse-grained categorical information, while the parameter gradients necessitate computationally intensive backward propagation processes.
More focus has been shed on exploring the features through efficient forward processes, as rich representations in feature spaces enable greater flexibility for more nuanced characterization of the InD-OoD disparity, which is also the main interest in this work.

In feature-based OoD detection, a powerful paradigm is feature {\it rectification}, which operates on the premise that extremely large or small feature activations are correlated with anomalous OoD predictions. 
Such extreme values in features are then clipped (rectified) in different ways to suppress OoD feature responses \cite{sun2021react,zhu2022boosting,xu2023vra,djurisicextremely,xuscaling,yuan2024discriminability,ahn2023line,song2022rankfeat}.
By contrast, another line of work aims at directly measuring the disparity between InD and OoD features through appropriate  {\it distances}, e.g., Mahalanobis \cite{lee2018simple,mullermahalanobis++} and $\ell_2$ distances \cite{sun2022out,park2023nearest} and other distance metrics \cite{zhang2022out,li2024characterizing}.
Despite their empirical success, the understandings on the InD-OoD disparity of these methods still remain limited.
In feature rectification, the clipping is conducted indiscriminately across all dimensions, which risks discarding useful feature responses for InD predictions.
For instance, globally removing the top-10\% of feature values over all dimensions is shown to cause nearly a 1.3\% accuracy drop on InD data \cite{sun2021react}.
In distance-based detection, it is assumed that the distance metrics can adequately capture the InD-OoD disparities in the feature space.
However, those distance metrics fail fundamentally when handling non-linear patterns commonly encountered in real-world data.
An example is presented in \cite{li2024characterizing} that the $\ell_2$ distance \cite{sun2022out} fails to distinguish InD and OoD given a non-linear data distribution of a Swiss Roll shape.
These findings underscore the importance of more nuanced interpretations on the InD-OoD disparity in the feature space.


In this work, we take the perspective of exploring the {\it non-linear feature subspace}, where InD data can be well described and yet OoD data cannot. 
To be specific, InD features from DNNs are commonly quite abstract and can intrinsically reside in a low-dimensional subspace, while the OoD features inherently cannot be well characterized by this subspace. 
Hence, such differences in this subspace can be an effective tool for distinguishing OoD data from InD data. 
Our perspective fundamentally differs from the aforementioned paradigms that directly operate in the whole feature space from DNNs.
Instead, our work focuses on investigating the intrinsic patterns of features in a subspace that well distinguish InD from OoD.
We further reveal that the immediate utilization of features or a naive linear subspace from InD data is insufficient in separating OoD data,
due to the non-linear characteristics of real-world feature distributions, as discussed in \cite{ndiour2020out,zaeemzadeh2021out,guan2023revisit}.
Nevertheless, the non-linear patterns that relate to the InD-OoD separability have not been fully exploited, and it still remains a significant and open question to identify a discriminative non-linear feature subspace for OoD detection.

In our method, the framework of Kernel Principal Component Analysis (KPCA) \cite{scholkopf1997kernel,scholkopf1998nonlinear} is leveraged to construct the non-linear feature subspace, where the InD-OoD disparities are sufficiently revealed.
Given a DNN $f(\boldsymbol{x}):\mathbb{R}^d\rightarrow\mathbb{R}^c$ well-trained on InD data, KPCA is applied to the penultimate layer features $\boldsymbol{z}\in\mathbb{R}^m$ of InD training samples and learns a non-linear subspace spanned by the principal components.
In inference, given features $\boldsymbol{\hat z}$ of an unknown sample $\boldsymbol{\hat x}$, one can obtain the reconstructed counterpart of $\boldsymbol{\hat z}$ by projecting $\boldsymbol{\hat z}$ to the non-linear subspace and re-projecting it back. 
Consequently, the reconstruction error between $\boldsymbol{\hat z}$ and its reconstructed counterpart can reflect the InD-OoD disparities and thereby serves as a detection metric. Here,
InD features are compactly allocated along the axes of principal components, leading to small reconstruction errors, while OoD features, which inherently are not well learned with this subspace,  give significantly larger reconstruction errors.

When applying KPCA to the OoD detection task, we emphasize two crucial issues w.r.t. efficacy and efficiency.
\begin{itemize}
    \item {\bf Non-linear kernel.} 
    By deploying KPCA, a non-linear mapping $\phi$ with its inducing kernel $k$, i.e., $k(\boldsymbol{z}_1,\boldsymbol{z}_2)=\langle\phi(\boldsymbol{z}_1),\phi(\boldsymbol{z}_2)\rangle$, is imposed on input features $\boldsymbol{z}$.
    To achieve effective OoD detection, the key is the identification of a suitable $k$ or $\phi$ that well models the non-linearity reflecting the InD-OoD disparities in the $\boldsymbol{z}$-space.
    \item {\bf Computational complexity.}
    KPCA is generally conducted on the kernel matrix ${\bf K}\in\mathbb{R}^{N_{\rm tr}\times N_{\rm tr}}$ of $N_{\rm tr}$ training samples.
    Obviously, when applied to large-scale detection scenarios with a huge InD data size $N_{\rm tr}$, such as $N_{\rm tr}=$ 1,281,167 of the ImageNet-1K \cite{deng2009imagenet} as InD, KPCA faces nearly prohibitive calculations and a significantly high computational complexity.
\end{itemize}

In this work, we formulate a KPCA OoD detection framework with the aforementioned two issues resolved.
Regarding the kernel selection, we reveal two vital non-linear patterns that closely relate to the InD-OoD disparities in the $\boldsymbol{z}$-space, and deduce a Cosine-Gaussian kernel that effectively characterizes such non-linearity.
KPCA w.r.t. this Cosine-Gaussian kernel promotes the separability between InD and OoD features and produces substantially improved discriminative reconstruction errors than PCA \cite{pearson1901liii,abdi2010principal}, as demonstrated in Figure \ref{fig:intro}.
On the other hand, to alleviate huge computations on the kernel matrix in the $\boldsymbol{z}$-space, we introduce explicit mappings $\Phi$ to approximate the Cosine-Gaussian kernel.
The approximated mapping $\Phi$ enables KPCA to learn principal components in the $\Phi(\boldsymbol{z})$-space and significantly reduces the computational complexity of reconstruction errors.
Moreover, in the context of OoD detection, $\Phi$ is also expected to benefit the differentiation between InD and OoD.
In this regard, we explore two representative mappings: the {\it data-independent} Random Fourier Features \cite{rahimi2007random} (RFFs) and the {\it data-dependent} Nystr\"om method \cite{NIPS2000_19de10ad}, and further refine the latter by involving task-specific insights on the InD data confidence.
Our refined Nystr\"om allows KPCA reconstruction errors to be even sensitive to distribution shifts and brings stronger detection results.

To summarize, this study further develops the core idea of KPCA-based reconstruction for OoD detection from our prior conference publication \cite{fang2024kpca} into a formalized and principled KPCA detection framework, with main contributions below:
\begin{itemize}
    \item[{\it (i)}]{\it On non-linear kernel selection.} 
    We reveal two non-linear patterns essential for InD-OoD separability: imbalanced feature norms and $\ell_2$ feature distances, in Section \ref{sec:non-linear-kernel}.
    These two properties not only serve as principled justifications for the Cosine-Gaussian kernel proposed in \cite{fang2024kpca}, but also provide crucial design guidelines for selecting and understanding kernel representations specific to differentiate InD and OoD.
    These insights are neither presented nor sufficiently analyzed in \cite{fang2024kpca}.
    \item[{\it (ii)}]{\it On task-driven kernel approximation.}
    We incorporate a data-dependent Nystr\"om approximation technique with a low-Energy sampling scheme specialized for the OoD detection task in Section \ref{sec:method:nys}.
    By reallocating landmarks toward boundary regions, this scheme actively promotes InD-OoD separability rather than simply preserving the global kernel reconstruction, yielding superior detection performance and reduced computational costs (Section \ref{sec:method:comp-complexity}).
    Together with the data-independent RFFs in \cite{fang2024kpca}, this work presents systematic explorations on efficient kernel approximation particularly tailored for OoD detection.
    Numerical approximation analyzes are provided in Section \ref{sec:method:approx}.
    \item[{\it (iii)}]{\it On learnable kernel representation.}
    We further explore a parametric realization of our KPCA OoD detection framework, as an extension of the non-parametric Cosine-Gaussian kernel in \cite{fang2024kpca}.
    This parametric learning paradigm supports learnable representations by optimizing kernel hyper-parameters and training neural networks with varied objectives, moving beyond fixed feature projections, as in Section \ref{sec:method:kernel-learning} and Appendix \ref{sec:app:learnable-kpca}.
    These primary and yet valuable explorations address the limitation noted in \cite{fang2024kpca} by showcasing data-driven kernel optimization for future work, and also further validate the flexibility of our framework.
    \item[{\it (iv)}]{\it Regarding empirical results}. 
    We  conduct extensive experiments in Section \ref{sec:exp} to validate the new State-Of-The-Art (SOTA) results and low computational demands of our KPCA detection framework, including comparisons with a broader variety of latest strong baselines and in-depth analyses for the two approximation methods.
\end{itemize}

In the remainder, the background knowledge is outlined in Section \ref{sec:background}.
Section \ref{sec:non-linear-kernel} presents the non-linear kernel selection in detail and Section \ref{sec:method:kernel-approximation} investigates our approximation techniques for efficient computations and explores learnable kernel representations.
Extensive empirical results are shown in Section \ref{sec:exp}.
Related works and concluding discussions are enclosed in Section \ref{sec:related-work} and Section \ref{sec:conclusion}, respectively.

\section{Background}
\label{sec:background}
\subsection{Out-of-Distribution Detection}
Generally, an OoD detection method justifies whether an input is from InD $\mathbb{P}_{\rm in}$ or OoD $\mathbb{P}_{\rm out}$ by a well-designed scoring function $S(\cdot)$ as follows:
\begin{equation}
\label{eq:ood-scoring}
\text{category\ of\ }\boldsymbol{x}=
\left\{
\begin{array}{ll}
\mathrm{InD},&S(\boldsymbol{x})\geq s,\\
\mathrm{OoD},&S(\boldsymbol{x})<s.
\end{array}
\right.
\end{equation}
The scoring function $S(\cdot)$ assigns a score $S(\boldsymbol{x})$ for an input $\boldsymbol{x}$.
If $S(\boldsymbol{x})$ is greater than a threshold $s$, $\boldsymbol{x}$ is viewed as an InD sample, and vice versa.
The threshold $s$ is usually selected such that most InD samples of the test dataset can be correctly identified.
Obviously, the detection performance depends on how well the scoring function $S(\cdot)$ characterizes the InD-OoD disparity.
Existing methods leverage different outputs from DNNs and design justified scores correspondingly; a thorough review can refer to Section \ref{sec:related-work}.

\subsection{PCA for Out-of-Distribution Detection}
The linear method, PCA \cite{pearson1901liii,abdi2010principal}, is considered for OoD detection \cite{guan2023revisit}.
For inputs $\boldsymbol{x}\in\mathbb{R}^d$, a well-trained DNN $f:\mathbb{R}^d\rightarrow\mathbb{R}^c$ learns features $\boldsymbol{z}\in\mathbb{R}^m$ before the last linear layer.
In the learning phase, PCA firstly calculates the centered covariance matrix ${\bf{\Sigma}}\in\mathbb{R}^{m\times m}$ on features $\{\boldsymbol{z}_i\}_{i=1}^{N_{\rm tr}}$ of the training (InD)  data $\{\boldsymbol{x}_i\}_{i=1}^{N_{\rm tr}}$:
\begin{equation}
\label{eq:covar-mat}
\boldsymbol{\Sigma}=\sum\nolimits_{i=1}^{N_{\rm tr}}\left(\boldsymbol{z}_i-\boldsymbol{\mu}_{\rm tr}
\right)\left(\boldsymbol{z}_i-\boldsymbol{\mu}_{\rm tr}\right)^\top,
\end{equation}
with $\boldsymbol{\mu}_{\rm tr}={1}/{N_{\rm tr}}\sum_{i=1}^{N_{\rm tr}}\boldsymbol{z}_i$. 
Through the eigendecomposition $\boldsymbol{\Sigma}={\bf{U}}{\bf{\Lambda}}{\bf{U}}^\top$, the projection matrix ${\bf{U}}_q\in\mathbb{R}^{m\times q}$ is obtained by taking the first $q$ columns of the eigenvector matrix ${\bf U}\in\mathbb{R}^{m\times m}$  w.r.t. the top-$q$  eigenvalues in  ${\bf \Lambda}\in\mathbb{R}^{m\times m}$.

In the inference phase, given a new sample $\boldsymbol{\hat x}\in\mathbb{R}^d$ and its feature $\boldsymbol{\hat z}\in\mathbb{R}^m$ from the DNN $f$, we can project the centered $\boldsymbol{\hat z}-\boldsymbol{\mu}_{\rm tr}$ to the ${\bf{U}}_q$-subspace and re-project it back.
The reconstruction error is computed as:
\begin{equation}
\label{eq:reconstruction-error}
e(\boldsymbol{\hat x})=\left\Vert{\bf{U}}_q{\bf{U}}_q^\top(\boldsymbol{\hat z}-\boldsymbol{\mu}_{\rm tr})-(\boldsymbol{\hat z}-\boldsymbol{\mu}_{\rm tr})\right\Vert_2.
\end{equation}
The subspace spanned by ${\bf{U}}_q$ is learned from InD data and thus gives high variances along principal components with informative patterns of InD, thereby yielding small reconstruction errors.  
In contrast, OoD data exhibits different patterns from InD data.
Thus, in the low-dimensional ${\bf{U}}_q$-subspace, OoD data is prone to produce large reconstruction errors along these extracted principal components.
The scoring function with PCA for OoD detection is thereby set as $S(\boldsymbol{\hat x})=-e(\boldsymbol{\hat x})$.

\subsection{Kernel Approximation}
\label{sec:background:rff-nys}

We outline the approximation techniques adopted in our KPCA detection framework.
In large-scale kernel methods, the expensive computations have been a long-standing problem, as an $N_{\rm tr}\times N_{\rm tr}$ kernel matrix $\bf K$ on training data requires $\mathcal{O}(N_{\rm tr}^2)$ kernel evaluations, $\mathcal{O}(N_{\rm tr}^2)$ space complexity, and $\mathcal{O}(N_{\rm tr}^3)$ time complexity in solving the optimization problem. 
To alleviate such computational burden, the data-independent random Fourier features \cite{rahimi2007random} and the data-dependent Nystr\"om method \cite{NIPS2000_19de10ad} are two mainstream methods.

\paragraph{Random Fourier features}
Random Fourier Features (RFFs) \cite{rahimi2007random} deploy explicit mappings with random features to approximate the kernel function $k$.
RFFs are based on Bochner's theorem \cite{rudin1962fourier} and construct random features $\phi_{\rm rff}$ for a shift-invariant kernel $k(\boldsymbol{z}_1,\boldsymbol{z}_2)=k(\boldsymbol{z}_1-\boldsymbol{z}_2)$:
\begin{equation}
\begin{aligned}
\label{eq:rff}
\phi_{\rm rff}(\boldsymbol{z})
&\triangleq\sqrt{\frac{2}{M_r}}\left[\phi_1(\boldsymbol{z}),\cdots,\phi_{M_r}(\boldsymbol{z})\right],\\
\phi_i(\boldsymbol{z})
&=\cos\ (\boldsymbol{z}^\top\boldsymbol{\omega}_i+u_i),i=1,\cdots,M_r,
\end{aligned}
\end{equation}
where $M_r$ denotes the number of RFFs, and $\boldsymbol{\omega}_i\in\mathbb{R}^m$ and $u_i\in\mathbb{R}$ are {\it i.i.d.} sampled from the Fourier transform of the kernel function $k(\cdot)$ and a uniform distribution $\mathcal{U}(0,2\pi)$, respectively.
We highlight two key properties of RFFs. \textit{(i)} Computing the explicit mapping $\phi_{\rm rff}(\boldsymbol{z})\in\mathbb{R}^{M_r}$, rather than the kernel matrix $\bf K$, can be significantly 
more efficient when $M_r\ll N_{\rm tr}$, while one could still guarantee  $k(\boldsymbol{z}_1,\boldsymbol{z}_2)\approx\phi_{\rm rff}(\boldsymbol{z}_1)^\top\phi_{\rm rff}(\boldsymbol{z}_2)$\cite{rahimi2007random}.
\textit{(ii)} The sampling distributions of $\boldsymbol{\omega}_i$ and $u_i$ are data-agnostic, and thereby RFFs are {\it data-independent}.

\paragraph{Nystr\"om method}
Instead of the explicit random features, the Nystr\"om method \cite{NIPS2000_19de10ad} presents low-rank approximation to the kernel matrix through finite sampling.
Firstly, $M_n$ {landmark} samples $\{\boldsymbol{\tilde z}_i\}_{i=1}^{M_n}$ are selected from the training set $\{\boldsymbol{z}_i\}_{i=1}^{N_{\rm tr}}$ ($M_n\ll N_{\rm tr}$), and then we can draw sub-matrices ${\bf\tilde K}\in\mathbb{R}^{{M_n}\times{M_n}}$ and ${\bf\bar K}\in\mathbb{R}^{{N_{\rm tr}}\times{M_n}}$ from the full kernel matrix $\bf K$ as ${\bf\tilde K}_{ij}=k(\boldsymbol{\tilde z}_i,\boldsymbol{\tilde z}_j)$ and ${\bf\bar K}_{ij}=k(\boldsymbol{z}_i,\boldsymbol{\tilde z}_j)$.
Nystr\"om method constructs a low-rank approximation of $\bf K$ as ${\bf\tilde K}_n=\bf\bar K\bf\tilde K^\dag\bf\bar K^\top$, where $^\dag$ denotes the matrix pseudo-inverse.
Obviously, Nystr\"om method is {\it data-dependent} and accelerates computations by working on the approximation associated with the selected $M_n$ landmarks.

\begin{figure*}[t]
\centering
\includegraphics[width=0.9\linewidth]{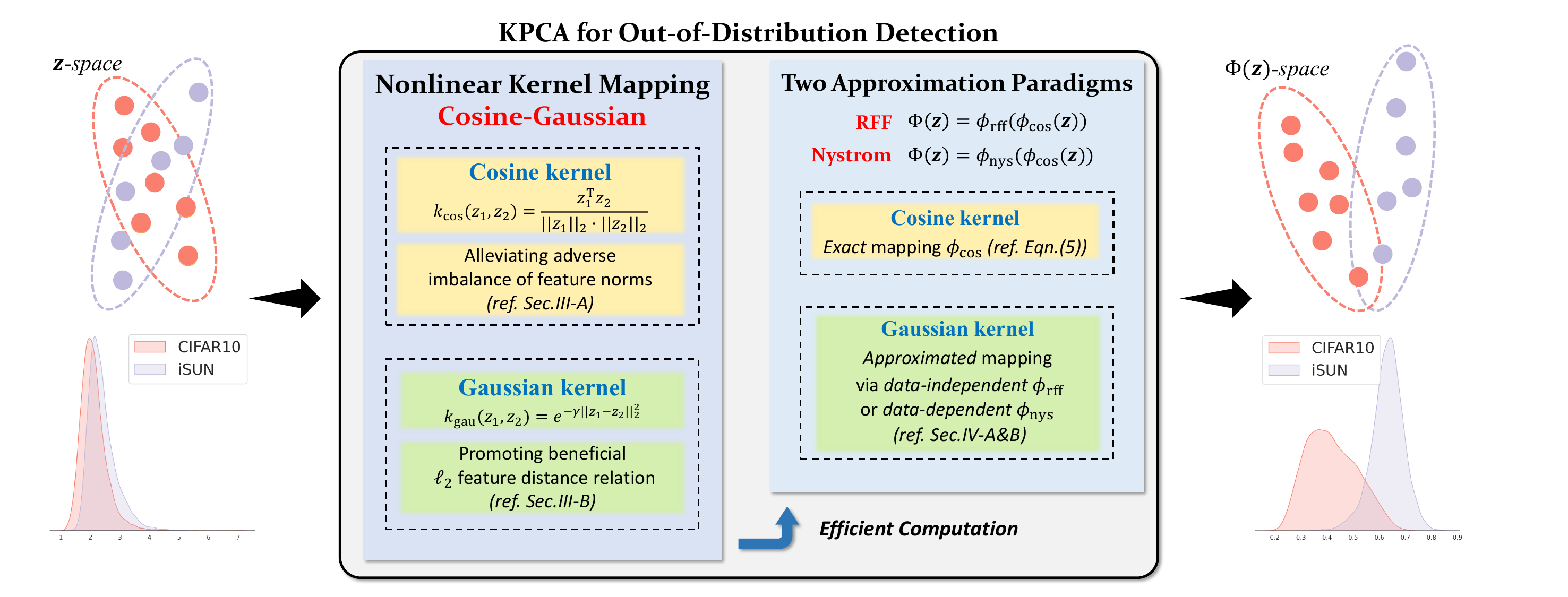}
\caption{
The framework of our KPCA detection method. 
A Cosine-Gaussian kernel is devised to model the non-linearity related to InD-OoD disparity in the $\boldsymbol{z}$-space (Section \ref{sec:non-linear-kernel}).
Explicit mappings $\Phi$ are built to approximate the Cosine-Gaussian kernel for efficient computations in the $\Phi(\boldsymbol{z})$-space (Section \ref{sec:method:kernel-approximation}).
The left and right histograms indicate the PCA and KPCA reconstruction errors on InD and OoD data, respectively, implying the effectiveness of the Cosine-Gaussian kernel in promoting the linear separability of between InD and OoD data.
}
\label{fig:kpca-framework}
\end{figure*}

\section{Non-Linear Kernel for OoD Detection}
\label{sec:non-linear-kernel}

In our work, we leverage Kernel PCA (KPCA) \cite{scholkopf1997kernel,scholkopf1998nonlinear} to identify discriminative non-linear feature subspaces for distinguishing InD and OoD data.
The key of KPCA lies in finding proper kernels to model the non-linearity that closely relates to InD-OoD separability in the $\boldsymbol{z}$-space, so as for discriminative non-linear principal components and distinguishable reconstruction errors on InD and OoD data.

Our KPCA detection framework is illustrated in Figure \ref{fig:kpca-framework}.
By analyzing InD and OoD features in the $\boldsymbol{z}$-space, we identify two non-linear patterns that reflect their disparities and derive a Cosine-Gaussian kernel to model such non-linearity: alleviating the adverse imbalance of feature norms through a Cosine kernel in Section \ref{sec:kernel:cosine} and promoting the beneficial feature distance relation by a Gaussian kernel in Section \ref{sec:kernel:gaussian}.

\subsection{Adverse Imbalance of Feature Norms}
\label{sec:kernel:cosine}

\begin{figure*}[t]
\centering
\includegraphics[width=0.99\linewidth]{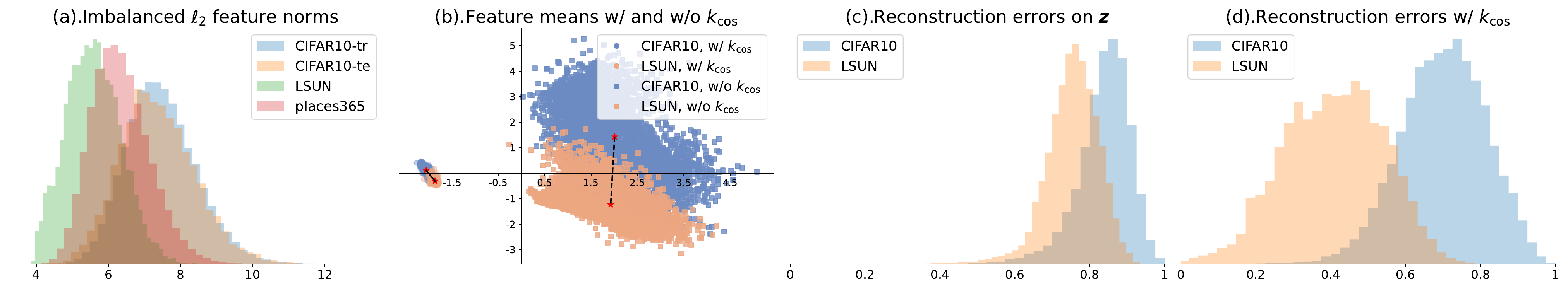}
\caption{Illustrations on the impact of the cosine kernel $k_{\rm cos}$. 
{\bf (a)} Density histograms on the imbalance of InD and OoD feature norms $\|\boldsymbol{z}\|_2$. 
{\bf (b)} The cosine kernel $k_{\rm cos}$ significantly reduces the distance between the InD feature means and the OoD feature means, which benefits the centering process in PCA. 
The red stars denote the feature means. The dashed and solid lines denote the distance between feature means without and with the cosine kernel $k_{\rm cos}$, respectively.
{\bf (c)\&(d)} The cosine kernel $k_{\rm cos}$ leads to more separable reconstruction errors of InD and OoD features and thereby boosts OoD detection performance.
InD: CIFAR10 \cite{krizhevsky2009learning}.
OoD: LSUN \cite{yu2015lsun} and places365 \cite{zhou2017places}.
}
\label{fig:exp-feat-norm}
\end{figure*}

Since InD and OoD data are assumed to be from different distributions, a substantial gap 
exists in their feature norms $\|\boldsymbol{z}\|_2$. 
As shown in Figure \ref{fig:exp-feat-norm}(a), samples in the CIFAR10 training and test datasets are  from InD, and thus have similar feature norms, while samples from OoD datasets exhibit significant deviations in feature norms.
In this subsection, we reveal that such norm imbalance between InD and OoD features has an adverse impact for OoD detection by violating the zero-mean assumption, when simply considering the linear subspace from PCA. 
Then, we further show that a Cosine kernel $k_{\rm cos}$ can effectively mitigate this issue with KPCA.

An essential pre-processing step of PCA is the {\it centering} on the inputs, i.e., inputs subtracted by their means in both learning and inference phases (ref. $(\boldsymbol{z}_i-\boldsymbol{\mu}_{\rm tr})$ in Eqn.\eqref{eq:covar-mat} and $(\boldsymbol{\hat z}-\boldsymbol{\mu}_{\rm tr})$ in Eqn.\eqref{eq:reconstruction-error}).
In the learning phase, centering is applied to the training data for computing the covariance matrix and the principal components.
Particularly, in inference, the centering is also required for given new inputs.
In OoD detection, the available feature mean for centering is the mean of InD training features, i.e., $\boldsymbol{\mu}_{\rm tr}$ in Eqn.\eqref{eq:covar-mat}, since the feature mean of OoD is unknown.
Meanwhile, note that the InD and OoD feature norms are imbalanced, which naturally induces a significant offset between their respective means, as shown by the right dashed line in Figure \ref{fig:exp-feat-norm}(b).
Therefore, centering OoD features using the InD feature means $\boldsymbol{\mu}_{\rm tr}$ is impractical for PCA, which results in misleading projections in the subspace and poor performance of reconstruction errors in Figure \ref{fig:exp-feat-norm}(c).

In this work, we note that a Cosine kernel $k_{\rm cos}$ defined below can well address this issue.
\begin{equation}
\begin{aligned}
\label{eq:kernel-cosine}
k_{\rm cos}(\boldsymbol{z}_1,\boldsymbol{z}_2)
&=\frac{\boldsymbol{z}_1^\top\boldsymbol{z}_2}{\|\boldsymbol{z}_1\|_2\cdot\|\boldsymbol{z}_2\|_2}
=\phi_{\rm cos}(\boldsymbol{z}_1)^\top\phi_{\rm cos}(\boldsymbol{z}_2),\\
\phi_{\rm cos}(\boldsymbol{z})&=\frac{\boldsymbol{z}}{\|\boldsymbol{z}\|_2}.
\end{aligned}
\end{equation}
$k_{\rm cos}$ is exactly induced from a non-linear mapping $\phi_{\rm cos}(\boldsymbol{z})$ that restricts inputs $\boldsymbol{z}$ onto a unit norm ball.
By this construction, given the InD and OoD features, $k_{\rm cos}$ inherently maps them to the unit sphere and thereby mitigates the adverse impact of their imbalanced feature norms.
In this way, the offset between InD and OoD feature means gets significantly reduced, as indicated by the left solid shorter line in Figure \ref{fig:exp-feat-norm}(b).
Therefore, the centering based on $\boldsymbol{\mu}_{\rm tr}$ w.r.t. the training data can now allocate OoD features nearby the same origin with InD features, so that the projection matrix learned from InD data can also be effective for OoD data, resulting in discriminative reconstruction errors in Figure \ref{fig:exp-feat-norm}(d).
In other words, a justified $\phi_{\rm cos}(\boldsymbol{z})$-space w.r.t. the cosine kernel $k_{\rm cos}$, instead of the original $\boldsymbol{z}$-space, ensures that the subspace spanned by the principal components can well characterize the InD-OoD disparity.

\subsection{Beneficial feature distance relation}
\label{sec:kernel:gaussian}

After the imbalance of InD and OoD feature norms gets alleviated through $k_{\rm cos}$, the resulting $\phi_{\rm cos}(\boldsymbol{z})$-space further enables explorations on another non-linear pattern related to the InD-OoD disparity, i.e., the beneficial $\ell_2$ feature distance relation, which could even promote the InD-OoD separability.

\begin{figure}[t]
\centering
\includegraphics[width=0.65\linewidth]{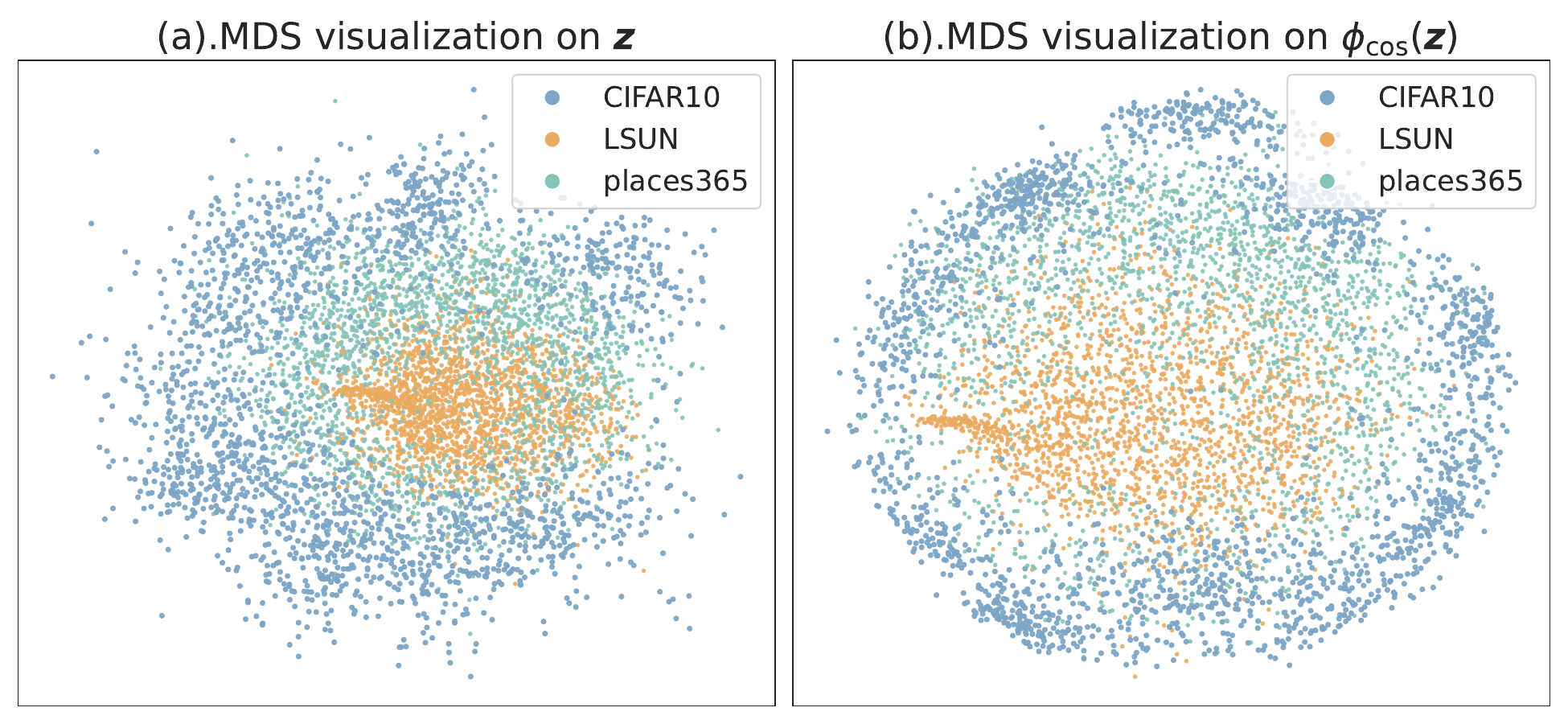}
\caption{Low dimensional embeddings via MDS \cite{borg2007modern} on InD and OoD data in the $\boldsymbol{z}$-space and $\phi_{\rm cos}(\boldsymbol{z})$-space, respectively.
InD: CIFAR10 \cite{krizhevsky2009learning}.
OoD: LSUN \cite{yu2015lsun} and places365 \cite{zhou2017places}.
}
\label{fig:exp-l2-distance}
\end{figure}

In Figure \ref{fig:exp-l2-distance}, Multi-Dimensional Scaling (MDS) \cite{borg2007modern} is applied to InD and OoD data with and without the cosine kernel $k_{\rm cos}$, respectively.
MDS seeks low-dimensional embeddings where the $\ell_2$ distance aligns well with that in the original space, making it an effective tool for analyzing the underlying $\ell_2$ distance structure of the data.
As illustrated in Figure \ref{fig:exp-l2-distance}, InD data in the $\phi_{\rm cos}(\boldsymbol{z})$-space is more concentrated w.r.t. the MDS embeddings built from the $\ell_2$ distance information.
Aside from the MDS visualizations, another work \cite{sun2022out} directly calculates the $\ell_2$ distances on $\ell_2$-normalized InD and OoD features as the detection score.
Empirical results in \cite{sun2022out} also indicate that the InD-InD $\ell_2$ distances are significantly smaller than the InD-OoD $\ell_2$ distances in the $\phi_{\rm cos}(\boldsymbol{z})$-space.
All these evidences suggest {\textit{(i)}} the insufficiency of the cosine kernel $k_{\rm cos}$ alone in capturing the non-linearity related to the InD-OoD disparity and {\textit{(ii)}} the promising potentials of leveraging the $\ell_2$ distance relation in the $\phi_{\rm cos}(\boldsymbol{z})$-space for separating InD and OoD data.

Note that a Gaussian kernel $k_{\rm gau}$, defined as
\begin{equation}
\label{eq:kernel-gau}
k_{\rm gau}(\boldsymbol{z}_1,\boldsymbol{z}_2)=e^{-\gamma\|\boldsymbol{z}_1-\boldsymbol{z}_2\|^2_2},
\end{equation}
is exactly able to preserve the $\ell_2$ distance information between samples.
Therefore, based on the above investigations, we leverage the Gaussian kernel $k_{\rm gau}$ on top of the cosine kernel $k_{\rm cos}$ to further model the discriminative $\ell_2$ feature distance existing in InD and OoD data in the $\phi_{\rm cos}(\boldsymbol{z})$-space.

To sum up, the eventual kernel we propose for the OoD detection task is a composite Cosine-Gaussian kernel.
This Cosine-Gaussian kernel models two vital non-linear properties hidden in InD and OoD features: alleviating the adverse imbalance of feature norms $\|\boldsymbol{z}\|_2$ and promoting the beneficial $\ell_2$ distance relation, and finally enables discriminative non-linear feature subspaces for separating InD and OoD data.

\section{Kernel Approximation for OoD Detection}
\label{sec:method:kernel-approximation}

Given the derived Cosine-Gaussian kernel for modeling the non-linearity, we even face challenges of the nearly prohibitive calculations on the kernel matrix ${\bf K}\in\mathbb{R}^{N_{\rm tr}\times N_{\rm tr}}$ under large-scale detection scenarios with a huge InD data size $N_{\rm tr}$, e.g., the common ImageNet-1K training set with $N_{\rm tr}=$1,281,167.
For efficient computations, we introduce approximation techniques in kernel methods, i.e., using an explicit non-linear mapping $\Phi$ to approximate the Cosine-Gaussian kernel, as illustrated in our framework of Figure \ref{fig:kpca-framework}.
In this way, KPCA learns non-linear principal components in the mapped $\Phi$-space, which excludes the kernel matrix $\bf K$ and leads to a significantly reduced computational complexity for calculating the reconstruction error as the detection score. 

In the composite Cosine-Gaussian kernel, the Cosine kernel $k_{\rm cos}$ directly admits an exact mapping $\phi_{\rm cos}$ in Eqn.\eqref{eq:kernel-cosine}.
In contrast, the  mapping associated with the Gaussian kernel $k_{\rm gau}$ is infinite-dimensional.
Nevertheless, $k_{\rm gau}$ can still be effectively approximated through finite-dimensional mappings.
Particularly, we explore two representative approximation mappings that differ in whether data is involved: \textit{(i)} data-independent random Fourier features \cite{rahimi2007random} in Section \ref{sec:method:rff}, and \textit{(ii)} data-dependent Nystr\"om approximation \cite{NIPS2000_19de10ad} specifically tailored for OoD detection in Section \ref{sec:method:nys}.
Discussions on the advantageous efficiency and analytical approximation properties are presented in Section \ref{sec:method:comp-complexity} and Section \ref{sec:method:approx}.

\subsection{Random Fourier Features}
\label{sec:method:rff}
The Random Fourier Features (RFFs) \cite{rahimi2007random} are leveraged to approximate the Gaussian kernel $k_{\rm gau}$.
As outlined in Section \ref{sec:background:rff-nys}, to build the approximated mapping $\phi_{\rm rff}$ in Eqn.\eqref{eq:rff}, $2M_r$ randomly-sampled weights $\boldsymbol{w}$ and $u$ are required.
To be specific, $M_r$ weights $u$ are sampled from a uniform distribution: $u\sim\mathcal{U}(0,2\pi)$, and the sampling distribution for $\boldsymbol{\omega}$ is the Fourier transform of the Gaussian kernel $k_{\rm gau}$, i.e., $\boldsymbol{\omega}\sim\mathcal{N}({\bf 0},\sqrt{2\gamma}{\bf{I}}_m)$, where $\gamma$ is the Gaussian kernel width.
Hence, $\phi_{\rm rff}$ in RFFs is agnostic of data and can be straightforwardly applied. 

Note that the Cosine-Gaussian kernel in our KPCA detection method is a composite of kernel functions.
The final non-linear mapping $\Phi$ is then formulated as $\Phi(\boldsymbol{z})\triangleq\phi_{\rm rff}(\phi_{\rm cos}(\boldsymbol{z}))$.
Given the mapping $\Phi$, in the learning phase, the training InD features $\{\boldsymbol{z}_i\}_{i=1}^{N_{\rm tr}}$ are mapped to $\Phi$-space, and a projection matrix ${\bf{U}}_q^\Phi$ is obtained by the eigendecomposition on the covariance matrix ${\bf{\Sigma}}^\Phi$ induced from $\{\Phi(\boldsymbol{z}_i)\}_{i=1}^{N_{\rm tr}}$.
In the inference phase, features $\boldsymbol{\hat z}$ of a new sample $\boldsymbol{\hat x}$ are firstly mapped to $\Phi(\boldsymbol{\hat z})$ and then compute the reconstruction error $e^\Phi$ based on ${\bf{U}}_q^\Phi$ and the mean $\boldsymbol{\mu}^\Phi_{\rm tr}$.
Algorithm \ref{alg:kernel-pca} outlines the key steps. 

\begin{algorithm}[t]
   \caption{The KPCA framework for OoD Detection}
   \label{alg:kernel-pca}
   \begin{algorithmic} [1]
   \STATE \COMMENT{\texttt{Learning phase: non-linear mapping}}
   \IF{\textsc{\underline{Random Fourier Features}}}
   \STATE Sample $\boldsymbol{\omega}_i\sim\mathcal{N}({\bf0},\sqrt{2\gamma}{\bf{I}}_m)$ and $u_i\sim\mathcal{U}({0},2\pi)$ for $i=1,\cdots,M_r$.
   \STATE Compute $\phi_i(\boldsymbol{z})=\cos\ (\boldsymbol{z}^\top\boldsymbol{\omega}_i+u_i)$ and $\phi_{\rm rff}(\boldsymbol{z})= \sqrt{\frac{2}{M_r}}\left[\phi_1(\boldsymbol{z}),\cdots,\phi_{M_r}(\boldsymbol{z})\right]$.
   \STATE Construct the mapping for Cosine-Gaussian kernel: $\Phi(\boldsymbol{z})\triangleq\phi_{\rm rff}(\phi_{\rm cos}(\boldsymbol{z}))$ with $\phi_{\rm cos}(\boldsymbol{z})=\frac{\boldsymbol{z}}{\|\boldsymbol{z}\|_2}$.
   
   \ELSIF{\textsc{\underline{Nystr\"om Approximation}}}
   \STATE Calculate energy of InD data $\{E_i=E(\boldsymbol{z}_i;f)\}_{i=1}^{N_{\rm tr}}$.
   \STATE Select $M_n$ landmarks $\{\boldsymbol{\tilde z}_i\}_{i=1}^{M_n}$ {w.r.t.} the smallest-$M_n$ energy in $\{E_i\}_{i=1}^{N_{\rm tr}}$.
    \STATE Compute the Cosine-Gaussian kernel matrix ${\bf\tilde K}$ w.r.t. the selected landmarks: ${\bf\tilde K}_{ij}=k_{\rm gau}(\phi_{\rm cos}(\boldsymbol{\tilde z}_i),\phi_{\rm cos}(\boldsymbol{\tilde z}_j))$.
   \STATE Apply eigendecomposition ${\bf\tilde K}={\bf\tilde U}{\bf\tilde\Lambda}{\bf\tilde U}^\top$ and construct the mapping $\Phi(\boldsymbol{z})\triangleq{\bf\tilde\Lambda}^{\text{-}\frac{1}{2}}{\bf\tilde U}^\top
   [\Phi_1(\boldsymbol{z}),\cdots,\Phi_{M_n}(\boldsymbol{z})]^\top$ with
   $\Phi_i(\boldsymbol{z})=
   k_{\rm gau}(\phi_{\rm cos}(\boldsymbol{ z}),\phi_{\rm cos}(\boldsymbol{\tilde z}_i))$.
   \ENDIF
   \vspace{0.1cm}
   \STATE \COMMENT{\texttt{Learning phase: construct subspace}}
   \STATE 
   Compute the training covariance matrix ${\bf{\Sigma}}^\Phi$ w.r.t. $\Phi$ using Eqn.\eqref{eq:covar-mat}, with the mean value
   $\boldsymbol{\mu}^\Phi_{\rm tr}=\frac{1}{N_{\rm tr}}\sum_{i=1}^{N_{\rm tr}}{\Phi(\boldsymbol{z}_i)}$.
   \STATE Apply eigendecomposition: ${\bf{\Sigma}}^\Phi={\bf{U}}^\Phi{\bf{\Lambda}}^\Phi{\bf{U}}^{\Phi\top}$ and take top-$q$ eigenvectors: ${\bf{U}}^\Phi_q={\bf{U}}^\Phi\left[:,:q\right]$.
   \ENSURE Approximated mapping $\Phi$, projection matrix ${\bf{U}}_q^\Phi$ and the mean value of InD features $\boldsymbol{\mu}^\Phi_{\rm tr}$.
   \vspace{0.2cm}
   \STATE \COMMENT{\texttt{Inference phase: reconstruction}}
   \STATE Given a new sample $\boldsymbol{\hat x}$ and its feature $\boldsymbol{\hat z}$ from the DNN, execute the non-linear mapping $\Phi(\boldsymbol{\hat z})$.
   \STATE Calculate the reconstruction error:\\
   $e^\Phi(\boldsymbol{\hat x})=\left\Vert{\bf{U}}^\Phi_q{\bf{U}}_q^{\Phi\top}({\Phi(\boldsymbol{\hat z})}-\boldsymbol{\mu}^\Phi_{\rm tr})-({\Phi(\boldsymbol{\hat z})}-\boldsymbol{\mu}^\Phi_{\rm tr})\right\Vert_2$.
   \ENSURE Detection score $S(\boldsymbol{\hat x})\triangleq-e^\Phi(\boldsymbol{\hat x})$.
\end{algorithmic}
\end{algorithm}

\subsection{Nystr\"om Approximation}
\label{sec:method:nys}

In this section, we elaborate how to leverage the Nystr\"om method \cite{NIPS2000_19de10ad} to construct an explicit mapping to approximate the Gaussian kernel for efficient OoD detection.
As outlined in Section \ref{sec:background:rff-nys}, Nystr\"om is commonly executed to build a low-rank approximation matrix ${\bf\tilde K}_n$ based on $M_n$ landmarks $\{\boldsymbol{\tilde z}_i\}_{i=1}^{M_n}$ selected from $N_{\rm tr}$ training samples $\{\boldsymbol{z}_i\}_{i=1}^{N_{\rm tr}}$.
This approximated ${\bf\tilde K}_n$ can further be decomposed as:
\begin{equation}
\begin{aligned}
\label{eq:nys}
[{\bf\tilde K}_n]_{ij}=\phi_{\rm nys}(\boldsymbol{z}_i)^\top\phi_{\rm nys}(\boldsymbol{z}_j),\quad
\phi_{\rm nys}(\boldsymbol{z})
\triangleq{\bf\tilde\Lambda}^{-1/2}{\bf\tilde U}^\top[k(\boldsymbol{z},\boldsymbol{\tilde z}_1),\cdots,k(\boldsymbol{z},\boldsymbol{\tilde z}_{M_n})]^\top.
\end{aligned}
\end{equation}
Note that ${\bf\tilde U}$ and ${\bf\tilde\Lambda}$ are obtained through the eigendecomposition on the submatrix ${\bf\tilde K}={\bf\tilde U}{\bf\tilde\Lambda}{\bf\tilde U}^\top$ with ${\bf\tilde K}_{ij}=k(\boldsymbol{\tilde z}_i,\boldsymbol{\tilde z}_j)$.
In this sense, the explicit mapping $\phi_{\rm nys}$ can be adopted as a data-driven approximation towards the Gaussian kernel $k_{\rm gau}$ through the selected $M_n$ landmarks.

Obviously, the sampling scheme for the landmarks directly determines the approximation performance of $\phi_{\rm nys}$, with studies ranging from naive uniform sampling to more sophisticated strategies \cite{JMLR:v6:drineas05a,kumar2012sampling}.
Nevertheless, in OoD detection contexts, the approximation performance of the Nystr\"om mapping is no longer the main concern; instead, we aim to identify a set of particular InD landmarks to build the approximated mapping, such that the mapped space w.r.t. these specific landmarks can help reflect the InD-OoD disparity and facilitate the subspace learning for effective OoD detection.

In this work, an {\it energy-based} sampling scheme is designed for Nystr\"om, tailored to boost OoD detection performance.
To be specific, we consider the following energy function $E$ applied to the logits of training samples $\{\boldsymbol{x}_i\}_{i=1}^{N_{\rm tr}}$ from the well-trained DNN $f$:
\begin{equation}
\label{eq:energy-func}
E(\boldsymbol{x};f)=E(\boldsymbol{z};f)=T\cdot\log\sum_{i=1}^{c}{\rm exp}\left(\frac{f_i(\boldsymbol{x})}{T}\right),
\end{equation}
where $f_i(\boldsymbol{x})$ denotes the $i$-th element of the logits $f(\boldsymbol{x})\in\mathbb{R}^c$ and $T$ indicates the temperature coefficient.
Here, high-energy samples correspond to those predictions, where the DNN $f$ assigns a dominant logit to one class, implying that those samples are likely from InD.
In contrast, low-energy samples demonstrate dispersed or nearly uniform logit distributions, indicating that they tend to align with OoD \cite{liu2020energy}.

\begin{figure}[t]
\centering
\includegraphics[width=0.6\linewidth]{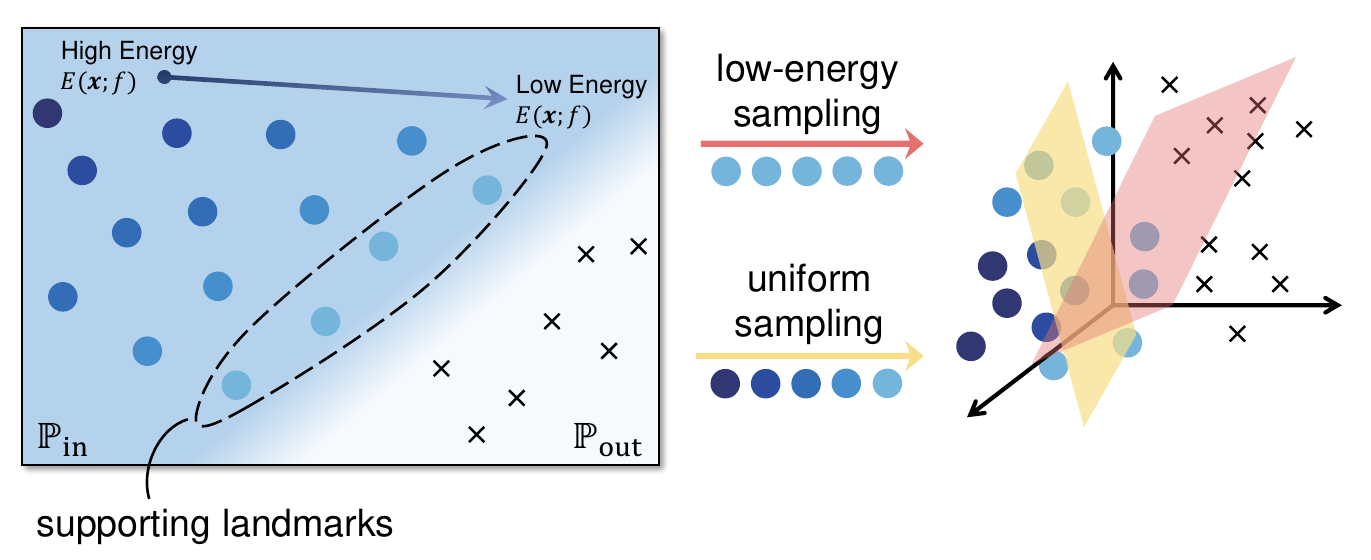}
\caption{An illustration on the comparisons between the naive uniform sampling and the proposed low-energy sampling, and their corresponding subspaces.
}
\label{fig:nystrom-subspace}
\end{figure}

The proposed energy-based sampling scheme selects $M_n$ {\it low-energy} InD training samples $\{\boldsymbol{\tilde x}_i\}_{i=1}^{M_n}$ w.r.t. the smallest energy values $E(\boldsymbol{\tilde x}_i;f)$ as landmarks to build the Nystr\"om approximation mapping $\phi_{\rm nys}$ of Eqn.\eqref{eq:nys}.
These landmarks are from InD and typically reside near the InD-OoD boundary, where $f$ is less confident on them.
For KPCA-based OoD detection, the detection threshold is governed by the largest gap in landmark coverage over the InD support, since the reconstruction error at any point can be controlled by its distance to nearby landmarks. 
Uniform or density-proportional sampling concentrates landmarks in high-density regions, leaving large coverage gaps near
low-density boundary areas, where the worst-case reconstruction error tends to be most pronounced.
Our low-Energy scheme reallocates landmarks toward these under-covered boundary regions, thereby shrinking the largest coverage gap and suppressing the detection threshold.
This yields a more compact acceptance region and directly limits false acceptance of OoD samples. 
We formalize this reasoning as follows, with an intuitive illustration in Figure \ref{fig:nystrom-subspace}.
\begin{proposition}
Let $\mathcal{Z}_{\rm in} \subset \mathbb{R}^m$ be the support of InD features and $e(\boldsymbol{z}; \tilde{\mathbf{Z}})$ be the KPCA reconstruction error derived from a landmark set $\tilde{\mathbf{Z}} = \{\tilde{\boldsymbol z}_i\}_{i=1}^{M_n}$. 
By reallocating landmarks from a global uniform distribution to the low-Energy regions of $\mathcal{Z}_{\rm in}$, the detection threshold $\gamma$ necessary to encompass the InD manifold can be effectively suppressed, yielding a more compact acceptance region $\mathcal{A} = \{\boldsymbol{z} \in \mathbb{R}^m \mid e(\boldsymbol{z}; \tilde{\mathbf{Z}}) \le \gamma\}$.
\end{proposition}

Based on the proposed low-energy supporting landmarks, the final non-linear mapping $\Phi$ via Nystr\"om approximation for the Cosine-Gaussian kernel is $\Phi(\boldsymbol{z})\triangleq\phi_{\rm nys}(\phi_{\rm cos}(\boldsymbol{z}))$.
The corresponding KPCA detection procedure is basically the same as KPCA via RFFs in Section \ref{sec:method:rff}, which is also outlined in Algorithm \ref{alg:kernel-pca}.
We use {\bf KPCA}$_{\rm rff}$ and {\bf KPCA}$_{\rm nys}$ to denote our KPCA detection method via RFFs and Nystr\"om, respectively.

\subsection{Computation Complexity} 
\label{sec:method:comp-complexity}

With approximation schemes, the computational efficiency of our KPCA detection framework is greatly improved.
As illustrated in Algorithm \ref{alg:kernel-pca}, 
the computation of the kernel matrix {\bf K} w.r.t.  $N_{\rm tr}$ training samples is skipped.
Particularly, in the inference phase, the KPCA reconstruction error $e^\Phi$ requires the approximated mapping $\Phi$, the projection matrix ${\bf{U}}^\Phi_q$ and the mean mapped training feature vector $\boldsymbol{\mu}^\Phi$.
Both ${\bf{U}}^\Phi_q$ and $\boldsymbol{\mu}^\Phi$
are computed only once in the learning phase and thus stored in preparation for inference.
Therefore, the computation cost of $\rm KPCA_{rff}$ and $\rm KPCA_{nys}$ in inference comes from the construction of the explicit mapping $\Phi$ on new features $\boldsymbol{\hat z}$.

Regarding $\Phi$, the primary mapping $\phi_{\rm cos}$ w.r.t. the common cosine kernel $k_{\rm cos}$ in $\rm KPCA_{rff}$ and $\rm KPCA_{nys}$ is an in-place operation and does not require additional computations.
Thus, the mappings $\phi_{\rm rff}$ and $\phi_{\rm nys}$ to approximate the Gaussian kernel $k_{\rm gau}$ take the main computation:
\begin{itemize}
    \item As illustrated in Eqn.\eqref{eq:rff} of $\phi_{\rm rff}$, $2M_r$ samplings of $\boldsymbol{\omega}_i$ and $u_i$, $M_r$ dot-products and $M_r$ additions are required for $\phi_{\rm RFF}(\boldsymbol{\hat z})$. Accordingly, the time and memory complexity of $\rm KPCA_{rff}$ is ${\cal O}(M_r)$.
    \item To build $\phi_{\rm nys}(\boldsymbol{\hat z})$ by Eqn.\eqref{eq:nys}, two eigenmatrices $\bf\tilde \Lambda$, $\bf\tilde U$ and $M_n$ kernel evaluations on the $M_n$ landmarks $\{\boldsymbol{\tilde z}_i\}_{i=1}^{M_n}$ are demanded. Hence, the time and memory complexity of $\rm KPCA_{nys}$ is ${\cal O}(M_n)$.
\end{itemize}

Note that there exists a relevant method KNN \cite{sun2022out} that also explores the imbalanced feature norms and the beneficial $\ell_2$ distance for OoD detection.
KNN takes a rather straightforward way and adopts the non-parametric $\ell_2$-distance nearest neighbor searching on $\ell_2$-normalized inputs as the score:
\begin{equation}
\label{eq:knn-score}
S_{\rm knn}(\boldsymbol{\hat x})=-\min_{i:1,\cdots,N_{\rm tr}}\left\|\frac{\boldsymbol{\hat z}}{\|\boldsymbol{\hat z}\|_2}-\frac{\boldsymbol{z}_i}{\|\boldsymbol{z}_i\|_2}\right\|_2.
\end{equation}
Aside from the distinctive kernel perspective, our KPCA further outperforms KNN in the computational overhead.
As in Eqn.\eqref{eq:knn-score}, KNN requires that all the training features $\{\boldsymbol{z}_i\}_{i=1}^{N_{\rm tr}}$ have to be stored at hand and iterated in inference, implying a heavy $\mathcal{O}(N_{\rm tr})$ time and memory complexity.
In contrast, the ${\cal O}(M_r)$/${\cal O}(M_n)$ of $\rm KPCA_{rff}$/$\rm KPCA_{nys}$ is substantially more efficient than the $\mathcal{O}(N_{\rm tr})$ of KNN given that $M_r,M_n\ll N_{\rm tr}$.
Detailed comparisons on detection performance and efficiency are provided in Sections \ref{sec:exp-main-comp} and \ref{sec:exp:comput-complex}.

\subsection{Analytical Discussion on Approximation Performance}
\label{sec:method:approx}
The primary objective of our approximated KPCA is to enhance OoD detection performance. 
In this section, we supplement an analysis of the approximation performance on InD and OoD data for a comprehensive investigation into our detection method.
While the statistical optimality of approximated KPCA via RFFs and Nystr\"om has been extensively studied within the traditional machine learning regime under well-defined assumptions \cite{lopez2014randomized,ullah2018streaming,sterge2020gain,hallgren2021kernel,sriperumbudur2022approximate,sterge2022statistical}, we extend the analysis to this deep learning regime, bridge theoretical approximation guarantees with practical detection scenarios, and provide actionable insights for implementation.

We consider the empirical version for the exact KPCA on a finite training set $\{\boldsymbol{z}_i\}_{i=1}^{N_{\rm tr}}$ without any approximations.
This requires that the reconstruction error should be calculated in the original $\boldsymbol{z}$-space based on the kernel matrix ${\bf K}\in\mathbb{R}^{N_{\rm tr}\times N_{\rm tr}}$.
To be specific, projections to the principal subspace based on $\bf K$ can be easily implemented \cite{scholkopf1997kernel,scholkopf1998nonlinear}, but how to reconstruct the projections in the principal subspace back to the original $\boldsymbol{z}$-space remains a non-trivial issue, known as the pre-image problem \cite{kwok2004pre}.
To address this issue, the following Proposition \ref{thm:reconstruction-error-kpca} \cite{fang2024kpca} shows a flexible approach to directly obtain the KPCA reconstruction error $e^{k}$ without building reconstructed features.

Given the eigen-decomposition on the kernel matrix $\bf K$ of the training set $\{\boldsymbol{z}_i\}_{i=1}^{N_{\rm tr}}$ as ${\bf K}={\bf U}^{k}{\bf \Lambda}^{k}{\bf U}^{{k}\top}$, and $\boldsymbol{k}_{\boldsymbol{\hat z}}=[k(\boldsymbol{z}_1,\boldsymbol{\hat z}),\cdots,k(\boldsymbol{z}_{N_{\rm tr}},\boldsymbol{\hat z})]^\top\in\mathbb{R}^{N_{\rm tr}}$ as a vector of kernel values between new data $\boldsymbol{\hat z}$ and $\{\boldsymbol{z}_i\}_{i=1}^{N_{\rm tr}}$, we have Proposition \ref{thm:reconstruction-error-kpca} below for exact KPCA reconstruction errors $e^k(\boldsymbol{\hat z})$ based on $\bf K$ (no approximation) with proofs in Appendix \ref{sec:app:proof}.
\begin{proposition}
\label{thm:reconstruction-error-kpca}
The exact KPCA reconstruction error $e^k(\boldsymbol{\hat z})$ can be calculated as:
\begin{equation}
\label{eq:reconstruction-error-kpca}
e^k(\boldsymbol{\hat z})=
\Vert
{\bf U}^{k\top}_p\boldsymbol{k}_{\boldsymbol{\hat z}}
\Vert_2,
\end{equation}
where ${\bf U}^{k}_p\in\mathbb{R}^{N_{\rm tr}\times p}$ includes the last $p$ columns of ${\bf U}^{k}$, i.e., those $p$ eigenvectors w.r.t. the smallest-$p$ eigenvalues in ${\bf \Lambda}^{k}$.
\end{proposition}

\begin{figure}[t]
\centering
\includegraphics[width=0.65\linewidth]{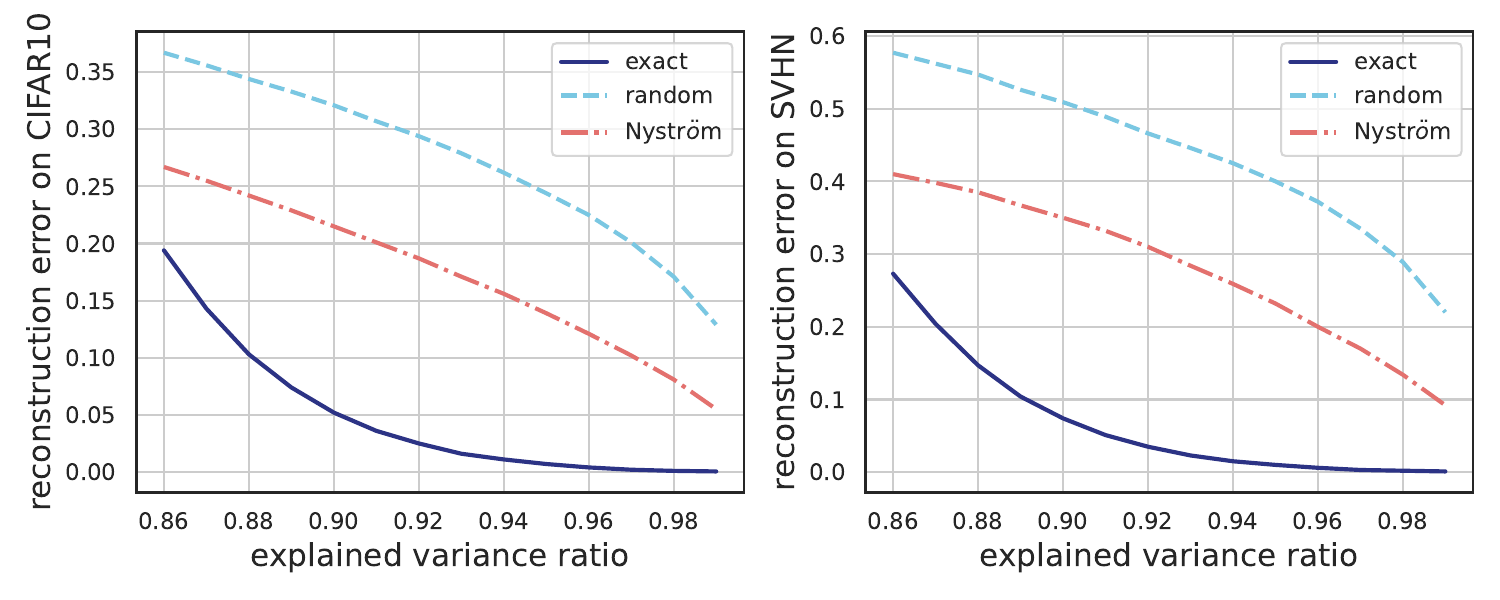}
\caption{Approximation performance of the reconstruction errors on InD (left) and OoD (right) data between exact KPCA and approximated KPCA based on RFFs (random) and Nystr\"om approximation (Nystr\"om).
}
\label{fig:theory-approx}
\end{figure}

Given Proposition \ref{thm:reconstruction-error-kpca}, numerical comparisons can be executed between exact KPCA and our approximated KPCA, with results of reconstruction errors on InD and OoD data shown in Figure \ref{fig:theory-approx}.
As illustrated, for this practical OoD detection scenario within the deep learning scheme, the RFFs and energy-based Nystr\"om lead to a marginal gap in the reconstruction errors when compared to exact KPCA, empirically verifying their approximation performance.
In addition, the data-dependent Nystr\"om achieves better convergence than the data-independent RFFs.

Again, we highlight that the approximation performance is not the focus of this study, and we recommend related literature \cite{lopez2014randomized,ullah2018streaming,sterge2020gain,hallgren2021kernel,sriperumbudur2022approximate,sterge2022statistical} for more in-depth and insightful theoretical explorations.
Instead, the essence of our work lies in \textit{(i)} exploring the non-linearity in InD and OoD data via task-specific kernels, and \textit{(ii)} facilitating OoD detection in practical large-scale scenarios via task-specific kernel approximations.

\subsection{Explorations on Learnable Kernels}
\label{sec:method:kernel-learning}
While the non-parametric Cosine-Gaussian kernel is validated with effective feature subspaces promoting InD-OoD separability, it would be interesting to further investigate learnable representations for optimizing non-linear feature subspaces.
Motivated by this, we consider data-driven kernel hyper-parameter optimization and parametric neural network formulations, each tailored to varied learning objectives.
These studies underscore an important insight: for learnable kernels, the objective function should be carefully designed to effectively capture distribution shifts, so that the representations are learned to be discriminative for InD and OoD in the non-linear low-dimensional feature subspace. 
Such primary investigations further demonstrate the potential of our framework for scenarios demanding higher flexibility and these tentative explorations may help inspire more comprehensive future work in this direction.
Details on implementation and numerical results are provided in Appendix \ref{sec:app:learnable-kpca}.

\section{Experiments}
\label{sec:exp}

\subsection{Setups}

\noindent{\bf Datasets and models.}\quad
In line with \cite{park2023nearest,song2024rankfeat}, the proposed KPCA detection framework is mainly evaluated on the large-scale ImageNet-1K benchmark \cite{deng2009imagenet} to illustrate its advantages in effectiveness and efficiency.
The ImageNet-1K dataset is more challenging and diverse than the classic CIFAR benchmark \cite{krizhevsky2009learning}: There are 1,281,167 training images of $224\times224\times3$ within 1,000 categories in ImageNet-1K, while CIFAR10 only contains 50,000 $32\times32\times3$ training images from 10 classes.
Besides, multiple different network structures are involved in experiments, including ResNet \cite{he2016deep}, MobileNet \cite{sandler2018mobilenetv2} and ViT (Vision Transformer) \cite{dosovitskiy2020vit} models.

For ImageNet-1K as InD, OoD datasets include iNaturalist \cite{van2018inaturalist}, SUN \cite{xiao2010sun}, Places \cite{zhou2017places} and Textures \cite{cimpoi2014describing}, which have been widely adopted in existing methods.
We select ResNet50 models \cite{he2016deep} of standard training and supervised contrastive learning \cite{khosla2020supervised}, MobileNetV2 \cite{sandler2018mobilenetv2} of standard training, and the prevalent ViT-B/16 \cite{dosovitskiy2020vit} for a fair comparison.
Images from ImageNet-1K and the paired OoD datasets are resized and cropped to $224\times224\times3$ as inputs for ResNet50 and MobileNetV2, and to $384\times384\times3$ for ViT-B/16, respectively.

\noindent{\bf Baselines.}\quad
$\rm KPCA_{rff}$ and $\rm KPCA_{nys}$ are compared with a wide variety of prevailing OoD detection methods, covering logits-based, gradients-based, and features-based ones.
A thorough review of these approaches can be found in Section \ref{sec:related-work}.

\noindent{\bf Metrics.}\quad
For the evaluation metrics, we employ the widely-used \textit{(i)} False Positive Rate (FPR) of OoD samples with 95\% true positive rate on the InD test dataset, and \textit{(ii)} Area Under the Receiver Operating Characteristic curve (AUROC).
The {\it average} FPR and AUROC values over the selected multiple OoD datasets are viewed as the final comparison metrics.

\noindent{\bf Implementation.}\quad
The source code of this work has been publicly released\footnote{\href{https://github.com/fanghenshaometeor/ood-kernel-pca-ext}{https://github.com/fanghenshaometeor/ood-kernel-pca-ext}}.
All experiments are executed on 1 NVIDIA GeForce RTX 3090 GPU.

Section \ref{sec:exp-main-comp} presents comparisons with these strong baselines on the ImageNet-1K benchmark with ResNet50 and ViT-B/16. 
Additional experiments, including those on MobileNetV2 and CIFAR, are provided in Appendix \ref{sec:app:exp-supp-ood} for reference, along with complete results w.r.t. the following Tables \ref{tab:exp-imgnet-r50-fuse}, \ref{tab:exp-varied-kernels}, \ref{tab:exp-ablation-sampling}.

\begin{table*}[ht]
    \centering
    \caption{Comparisons among a wide variety of OoD detection methods with \textbf{ResNet50} models trained on \textbf{ImageNet-1K}.}
    \resizebox{\textwidth}{!}{
    \begin{tabular}{c|cc cc cc cc|cc}
    \toprule
    \multirow{3}{*}{method} & \multicolumn{8}{c|}{OoD data sets} & \multicolumn{2}{c}{\multirow{2}{*}{\bf AVERAGE}} \\
    & \multicolumn{2}{c}{iNaturalist} & \multicolumn{2}{c}{SUN} & \multicolumn{2}{c}{Places} & \multicolumn{2}{c|}{Textures} &&\\
    & FPR$\downarrow$ & AUROC$\uparrow$ & FPR$\downarrow$ & AUROC$\uparrow$ &
    FPR$\downarrow$ & AUROC$\uparrow$ & FPR$\downarrow$ & AUROC$\uparrow$ & FPR$\downarrow$ & AUROC$\uparrow$ \\
    \midrule
    \multicolumn{11}{c}{\bf Standard Training}\\
    MSP \cite{hendrycks2016baseline} & 54.99 & 87.74 & 70.83 & 80.86 & 73.99 & 79.76 & 68.00 & 79.61 & 66.95 & 81.99 \\
    MaxLogit \cite{hendrycks2022scaling} & 54.49 & 91.05 & 65.45 & 84.96 & 67.60 & 83.69 & 57.09 & 86.71 & 61.16 & 86.60 \\ 
    ODIN \cite{liang2018enhancing} & 47.66 & 89.66 & 60.15 & 84.59 & 67.89 & 81.78 & 50.23 & 85.62 & 56.48 & 85.41 \\
    Energy \cite{liu2020energy} & 55.72 & {89.95} & {59.26} & {85.89} & 64.92 & 82.86 & 53.72 & 85.99 & 58.41 & 86.17 \\
    Mahala \cite{lee2018simple} & 97.00 & 52.65 & 98.50 & 42.41 & 98.40 & 41.79 & 55.80 & 85.01 & 87.43 & 55.47 \\
    ViM \cite{wang2022vim} & 68.86 & 87.13 & 79.62 & 81.67 & 83.81 & 77.80 & 14.95 & 96.74 & 61.81 & 85.83 \\
    DML \cite{zhang2023decoupling} & 47.32 & 91.61 & 57.40 & 86.14 & 61.43 & 84.68 & 52.80 & 86.72 & 54.74 & 87.28 \\
    KNN \cite{sun2022out} & 59.00 & 86.47 & 68.82 & 80.72 & 76.28 & 75.76 & 11.77 & 97.07 & 53.97 & 85.01 \\
    SHE \cite{zhang2022out} & 34.22 & 90.18 & 54.19 & 84.69 & {\bf45.35} & {\bf90.15} & 45.09 & 87.93 & 44.71 & 88.24 \\
    GradNorm \cite{huang2021importance} & {\bf31.24} & {\bf91.79} & {\bf38.53} & 88.87 & 46.29 & 86.28 & 46.76 & 83.66 & 40.70 & 87.63 \\
    \rowcolor{tabgray}{\bf KPCA}$_{\rm rff}$ & {50.07} & 89.32 & 62.56 & 83.74 & 72.76 & 78.91 &	{\bf9.02} & {\bf98.14} & 48.60 & 87.53 \\
    \rowcolor{tabgray}{\bf KPCA}$_{\rm nys}$ & 38.61 & 91.18 & 43.47 & {\bf90.20} & 57.35 & 85.30 & 13.16 & 97.41 & {\bf38.15} & {\bf91.02} \\
    \midrule
    \multicolumn{11}{c}{\bf Regularized Training}\\
    SSD+ \cite{sehwag2020ssd} & 57.16 & 87.77 & 78.23 & 73.10 & 81.19 & 70.97 & 36.37 & 88.52 & 63.24 & 80.09 \\
    KNN+ \cite{sun2022out} & 30.18 & 94.89 & 48.99 & 88.63 & 59.15 & 84.71 & 15.55 & 95.40 & 38.47 & 90.91 \\
    TNN \cite{li2024characterizing} & 28.73 & 95.08 & 45.38 & 88.82 & 56.54 & 85.05 & 18.72 & 94.25 & 37.34 & 90.80 \\
    Mahala++ \cite{mullermahalanobis++} & 32.47 & 95.16 & 63.65 & 86.57 & 69.19 & 84.01 & 14.77 & 95.61 & 45.02 & 90.34 \\
    MCF \cite{zhang2023decoupling} & 36.29 & 93.77 & 51.18 & 89.50 & 57.38 & 86.78 & 28.46 & 94.35 & 43.33 & 91.10 \\
    MNC \cite{zhang2023decoupling} & 10.94 & 97.88 & 25.34 & 94.49 & 34.99 & 91.82 & 50.57 & 85.21 & 30.46 & 92.35 \\
    DML+ \cite{zhang2023decoupling} & 13.57 & 97.50 & 30.21 & 94.01 & 39.06 & 91.42 & 36.31 & 89.70 & 29.79 & 93.16 \\
    NNGuide \cite{park2023nearest} & 12.02 & 97.47 & 31.62 & 91.66 & 38.88 & 90.12 & 24.93 & 91.52 & 26.86 & 92.69 \\
    \rowcolor{tabgray}{\bf KPCA}$_{\rm rff}$ & {23.61} & {95.86} & {41.07} & {91.25} & {53.52} & {87.27} & {10.23} & {97.04} & {32.11} & {92.86} \\
    \rowcolor{tabgray}{\bf KPCA}$_{\rm nys}$ & {\bf6.88} & {\bf98.43} & {\bf19.22} & {\bf94.84} & {\bf29.79} & {\bf92.31} & 24.49 & 94.19 & {\bf20.09} & {\bf94.94} \\
    \bottomrule
    \end{tabular}}
    \label{tab:exp-imgnet-r50}
\end{table*}

\subsection{Comparisons on Detection Performance}
\label{sec:exp-main-comp}

\subsubsection{Results on ResNet}
Table \ref{tab:exp-imgnet-r50} shows the detection performance of various detection methods on ResNet50 models of standard training and regularized training.
For the former, our KPCA employs the PyTorch-released ResNet50 checkpoint \cite{paszke2019pytorch} that is trained by a standard cross-entropy loss.
For the latter, our KPCA adopts the ResNet50 checkpoint released with KNN \cite{sun2022out}, which is trained with supervised contrastive learning \cite{khosla2020supervised}.
In both cases, the detection score of KPCA is the negative of the reconstruction error $S(\boldsymbol{\hat x})=-e^\Phi(\boldsymbol{\hat x})$.

In Table \ref{tab:exp-imgnet-r50},  our KPCA detection method achieves the lowest FPR and highest AUROC values, and indicates that exploring non-linear patterns is more advantageous than other compared methods in distinguishing OoD data from InD data.
Besides, the substantial improvements of $\rm KPCA_{nys}$ over $\rm KPCA_{rff}$ indicate that the data-dependent Nystr\"om approximation is more beneficial in learning a discriminative InD subspace than the data-independent RFFs.

\subsubsection{Results on ViT}
Table \ref{tab:exp-imgnet-vit} shows comparisons on the ViT-B/16 model. We follow the settings in \cite{park2023nearest} and adopt a ViT-B/16 checkpoint that is sequentially pre-trained on ImageNet-21K and fine-tuned on ImageNet-1K.
This transfer learning is a common and practical learning scheme.
As shown in Table \ref{tab:exp-imgnet-vit}, $\rm KPCA_{rff}$ and $\rm KPCA_{nys}$ are suitable for features learned by the visual attention mechanism of ViT and achieve superior performance than KNN \cite{sun2022out}, indicating the effectiveness of appropriate non-linear kernels.
Besides, the ensemble of $\rm KPCA_{rff}$ and $\rm KPCA_{nys}$ achieves new SOTA detection performance over other compared baselines.

\begin{table*}[ht]
    \centering
    \caption{Comparisons among a wide variety of OoD detection methods with \textbf{ViT-B/16} trained on \textbf{ImageNet-1K}.}
    \resizebox{\textwidth}{!}{
    \begin{tabular}{c|cc cc cc cc|cc}
    \toprule
    \multirow{3}{*}{method} & \multicolumn{8}{c|}{OoD data sets} & \multicolumn{2}{c}{\multirow{2}{*}{\bf AVERAGE}} \\
    & \multicolumn{2}{c}{iNaturalist} & \multicolumn{2}{c}{SUN} & \multicolumn{2}{c}{Places} & \multicolumn{2}{c|}{Textures} &&\\
    & FPR$\downarrow$ & AUROC$\uparrow$ & FPR$\downarrow$ & AUROC$\uparrow$ &
    FPR$\downarrow$ & AUROC$\uparrow$ & FPR$\downarrow$ & AUROC$\uparrow$ & FPR$\downarrow$ & AUROC$\uparrow$ \\
    \midrule
    MSP \cite{hendrycks2016baseline} & 27.58 & 93.96 & 57.43 & 85.18 & 61.13 & 84.34 & 53.21 & 84.96 & 49.84 & 87.11 \\
    Energy \cite{liu2020energy} & 12.64 & 97.34 & 48.05 & 86.47 & 56.41 & 82.22 & 46.79 & 85.72 & 40.97 & 87.94 \\
    Mahala \cite{lee2018simple} & 4.94 & 98.85 & 58.77 & 88.15 & 65.62 & 85.46 & 43.49 & 90.30 & 43.21 & 90.69 \\
    ViM \cite{wang2022vim} & 3.42 & 99.20 & 49.66 & 88.05 & 59.69 & 83.68 & 42.55 & 88.46 & 38.83 & 89.85 \\
    GradNorm \cite{huang2021importance} & 14.06 & 96.62 & 46.68 & 86.60 & 56.70 & 82.84 & 43.37 & 87.73 & 40.20 & 88.45 \\
    ReAct \cite{sun2021react} & 15.70 & 96.81 & 53.23 & 84.87 & 61.68 & 80.08 & 52.73 & 83.90 & 45.84 & 86.41 \\
    KNN \cite{sun2022out} & 29.46 & 94.07 & 72.15 & 83.88 & 74.17 & 81.47 & 51.21 & 87.18 & 56.75 & 86.65 \\
    MaxLogit \cite{hendrycks2022scaling} & 13.19 & 97.19 & 47.45 & 86.80 & 54.36 & 83.13 & 44.70 & 86.11 & 39.93 & 88.31 \\ 
    NNGuide \cite{park2023nearest} & 9.17 & 97.96 & 45.64 & 90.03 & 53.82 & {\bf87.25} & {\bf39.26} & {\bf90.01} & 36.97 & {\bf91.31} \\
    Mahala++ \cite{mullermahalanobis++} & 4.69 & 98.92 & 54.68 & 88.86 & 64.09 & 85.69 & 37.64 & 91.63 & 40.28 & 91.28 \\
    RankFeat+Weight \cite{song2022rankfeat,song2024rankfeat} & 12.14 & 97.48 & 48.69 & 86.14 & 55.80 & 82.22 & 45.94 & 85.82 & 40.64 & 87.91 \\
    \rowcolor{tabgray}{\bf KPCA}$_{\rm rff}$ & 10.30 & 97.88 & 68.73 & 83.47 & 76.59 & 78.86 & 49.40 & 89.15 & 51.25 & 87.34 \\
    \rowcolor{tabgray}{\bf KPCA}$_{\rm nys}$ & 4.20 & 99.00 & 49.30 & 87.15 & 61.00 & 82.47 & 53.17 & 87.04 & 41.92 & 88.92 \\
    \rowcolor{tabgray}{\bf KPCA}$_{\rm rff}$+{\bf KPCA}$_{\rm nys}$ & {\bf1.06} & {\bf99.65} & {\bf30.07} & {\bf90.52} & {\bf41.30} & {86.11} & 56.38 & 78.08 & {\bf32.20} & 88.59 \\
    \bottomrule
    \end{tabular}}
    \label{tab:exp-imgnet-vit}
\end{table*}

\subsubsection{Comparisons with feature rectification}
A series of methods propose that the abnormal responses in features lead to over-confidence on OoD data and should be removed.
Generally, those extremely large or small feature values are viewed as anomalous information causing OoD predictions.
Therefore, a range of techniques have been developed to clip extreme feature values \cite{sun2021react,song2022rankfeat,zhu2022boosting,xu2023vra,djurisicextremely,xuscaling,ahn2023line,yuan2024discriminability}, termed feature rectification.
Feature rectification still relies on an additional detector applied to the refined features or logits to produce detection scores, yet it has demonstrated outstanding detection performance and also a highly-flexible compatibility with other features-based detection methods \cite{sun2022out,guan2023revisit}.

Our KPCA detection method can be equipped with feature rectification for even improved detection results.
To be specific, a fusion strategy is proposed in \cite{guan2023revisit} to combine regularized PCA reconstruction errors with feature rectification.
Thereby, for a fair comparison, we follow this fusion strategy in \cite{guan2023revisit} and boost KPCA reconstruction errors via a pioneering, simple, and effective feature rectification method ReAct \cite{sun2021react}.
Table \ref{tab:exp-imgnet-r50-fuse} presents comparisons with more feature rectification methods on ImageNet-1K with ResNet50.

\begin{table}[t]
    \centering
    \caption{Detection performance of KPCA enhanced by feature rectification. Results are of {\bf ResNet50} on {\bf ImageNet-1K}.}
    \begin{tabular}{l|cc}
    \toprule
    \multirow{2}{*}{method} & \multicolumn{2}{c}{\bf AVERAGE} \\
    & FPR$\downarrow$ & AUROC$\uparrow$ \\
    \midrule
    DICE \cite{sun2022dice} & 35.71 & 90.92 \\
    ReAct \cite{sun2021react} & 31.43 & 92.95\\
    BATS \cite{zhu2022boosting} & 27.11 & 94.28 \\
    VRA \cite{xu2023vra} & 25.49 & 94.57 \\
    ASH-B \cite{djurisicextremely} & 22.73 & 95.06 \\
    SCALE \cite{xuscaling} & 20.05 & 95.71 \\
    LINe \cite{ahn2023line} & 20.69 & 95.03 \\
    DDCS \cite{yuan2024discriminability} & 19.36 & 95.55 \\
    GradOrth\cite{behpour2023gradorth} & 18.57 & 96.31 \\
    DICE \cite{sun2022dice}+ReAct & 28.57 & 93.30 \\
    GradNorm\cite{huang2021importance}+ReAct & 25.13 & 94.22 \\
    LowDim\cite{wu2024low}+ReAct & 23.03 & 95.45 \\
    NNGuide \cite{park2023nearest}+ReAct & 19.72 & 95.45 \\
    RegPCA\cite{guan2023revisit}+ReAct & 18.66 & 95.76 \\
    \rowcolor{tabgray}{\bf KPCA}$_{\rm rff}$+ReAct & 17.68 & 95.98 \\
    \rowcolor{tabgray}{\bf KPCA}$_{\rm nys}$+ReAct & {\bf15.26} & {\bf96.79} \\
    \bottomrule
    \end{tabular}
    \label{tab:exp-imgnet-r50-fuse}
\end{table}

\begin{table}[t]
    \centering
    \caption{Comparisons on the computation complexity among KNN, $\rm KPCA_{rff}$ and $\rm KPCA_{nys}$ ({\bf ResNet50} on {\bf ImageNet-1K}).
    Experiments are executed on the same machine for a fair comparison. 
    KNN is implemented via Faiss \cite{johnson2019billion}.}
    \begin{tabular}{c|c|cc}
    \toprule
    method & \makecell[c]{time and memory\\complexity} & time cost (ms) & storage cost \\
    \midrule
    KNN & \makecell[c]{$\mathcal{O}(N_{\rm tr})$\\$N_{\rm tr}=1,281,167$} & $\approx$ 15.59 & $\approx$ 20 GiB \\
    \rowcolor{tabgray}{\bf KPCA}$_{\rm rff}$ & $\mathcal{O}(M_r)$, $M_r=4,096$ & $\approx$ 0.464  & $\approx$ 93 MiB \\
    \rowcolor{tabgray}{\bf KPCA}$_{\rm nys}$ & $\mathcal{O}(M_n)$, $M_n=2,048$ & $\approx$ 0.212 & $\approx$ 83 MiB\\
    \bottomrule
    \end{tabular}
    \label{tab:exp-complexity}
\end{table}

Table \ref{tab:exp-imgnet-r50-fuse} indicates the superior detection performance of KPCA boosted by feature rectification. 
Our KPCA equipped with ReAct outperforms other powerful feature rectification methods \cite{zhu2022boosting,xu2023vra,djurisicextremely,xuscaling,ahn2023line,yuan2024discriminability}.
Besides, under the same feature-clipping technique, KPCA also achieves the lowest FPR and highest AUROC values among other strong baselines \cite{sun2022dice,huang2021importance,wu2024low,park2023nearest,guan2023revisit}.
The potential of KPCA can be further enhanced by incorporating more advanced feature rectification techniques.

All these experiments support the critical role of the non-linearity for separating InD and OoD data.
The derived Cosine-Gaussian kernel and the corresponding non-linear feature subspace well seize the InD-OoD disparities and produce reconstruction errors that are sensitive to distribution shifts with new SOTA detection results.

\subsection{Comparisons on Computation Complexity}
\label{sec:exp:comput-complex}

We provide numerical results on the time and memory complexity of our KPCA detection method in the inference phase, i.e., the computation cost for calculating the KPCA reconstruction error $e^\Phi$ given features $\boldsymbol{\hat z}$ of a new sample $\boldsymbol{\hat x}$.
Particularly, the KNN method \cite{sun2022out} is incorporated as a reference baseline, since both the Cosine-Gaussian kernel and the nearest neighbor searching explore the imbalanced feature norms and the $\ell_2$ distance relation for OoD detection but in rather different perspectives and implementations.

Table \ref{tab:exp-complexity} shows detailed numerical comparisons on ImageNet-1K with ResNet50.
In the inference phase for new features $\boldsymbol{\hat z}$, it is essential for KNN \cite{sun2022out} to store and iterate all the 1,281,167 training features for the nearest neighbor searching, costing nearly 20 GiB memory and 16 ms to calculate the nearest distance for $\boldsymbol{\hat z}$.
In contrast, as outlined in Algorithm \ref{alg:kernel-pca}, our KPCA constructs explicit mappings without keeping and visiting the whole training set in inference, and achieves a significantly cheap computation complexity:
\begin{itemize}
    \item $\rm KPCA_{rff}$ directly samples $M_r=4,096$ weights of $\boldsymbol{\omega}_i$ and $u_i$ to obtain mapped features $\Phi(\boldsymbol{\hat z})$, and adopts pre-calculated projection matrix ${\bf{U}}^\Phi_q$ and mean training vector $\boldsymbol{\mu}^\Phi$ to determine the reconstruction error $e^\Phi$.
    \item $\rm KPCA_{nys}$ only needs to keep and iterate $M_n=2,048$ support landmarks $\{\boldsymbol{\tilde z}\}_{i=1}^{M_{n}}$ and pre-calculated matrices $\bf\tilde\Lambda$, $\bf\tilde U$, ${\bf{U}}^\Phi_q$ and $\boldsymbol{\mu}^\Phi$ to acquire $\Phi(\boldsymbol{\hat z})$ and $e^\Phi$.
\end{itemize}
As shown in Table \ref{tab:exp-complexity}, both $\rm KPCA_{rff}$ and $\rm KPCA_{nys}$ are highly resource-efficient compared with KNN, and complete inference in less than 1 ms and fewer than 100 MiB of memory occupancy.
Besides, $\rm KPCA_{nys}$ even outperforms $\rm KPCA_{rff}$, indicating the superiority of the data-dependent approximation over data-independent random features.

\begin{table}[t]
    \centering
    \caption{Detection results among a variety of kernels approximated via random features and the Nystr\"om approximation.}
    \begin{tabular}{ll|cc}
    \toprule
    \multirow{2}{*}{kernel} & \multirow{2}{*}{approx.} & \multicolumn{2}{c}{\bf AVERAGE} \\
    & & FPR$\downarrow$ & AUROC$\uparrow$ \\
    \midrule
    PCA & - & 84.41 & 57.46 \\
    \midrule
    Cosine & - & 58.40 & 83.86 \\
    \multirow{2}{*}{Polynomial} & random & 96.91 & 47.22 \\
    & Nystr\"om & 86.33 & 64.65 \\
    \multirow{2}{*}{Laplacian} & random & 94.82 & 50.17 \\
    & Nystr\"om & 81.41 & 38.72 \\
    \multirow{2}{*}{Gaussian} & random & 94.88 & 50.60 \\
    & Nystr\"om & 38.15 & 35.44 \\
    \midrule
    \multirow{2}{*}{Cosine-Poly.} & random & 68.01 & 77.95 \\
    & Nystr\"om & 58.42 & 83.79 \\
    \multirow{2}{*}{Cosine-Lap.} & random & 62.54 & 80.60 \\
    & Nystr\"om & 54.05 & 86.71 \\
    \multirow{2}{*}{Cosine-Gau.} & random & 48.60 & 87.53 \\ 
    & Nystr\"om & 38.15 & 91.02 \\
    \bottomrule
    \end{tabular}
    \label{tab:exp-varied-kernels}
\end{table}

\subsection{Alternative Kernels}
\label{sec:exp:alter-kernels}

We explore alternative kernels for OoD detection, extending beyond the proposed Cosine-Gaussian kernel.
\begin{itemize}
    \item A Laplacian kernel $k_{\rm lap}$ characterizes the $\ell_1$ distance between two samples $\boldsymbol{z}_1$ and $\boldsymbol{z}_2$:
    \begin{equation}
    \label{eq:kernel-lap}
    k_{\rm lap}(\boldsymbol{z}_1,\boldsymbol{z}_2)=e^{-\gamma\|\boldsymbol{z}_1-\boldsymbol{z}_2\|_1}.
    \end{equation}
    The Laplacian kernel $k_{\rm lap}$ is shift-invariant, thus RFFs of Eqn.\eqref{eq:rff} can be applied to approximate $k_{\rm lap}$.
    Note that the Fourier transform of $k_{\rm lap}$ to sample $\boldsymbol{\omega}_i$ is a Cauchy distribution $\boldsymbol{\omega}\sim\frac{\gamma^2}{\pi\gamma(\boldsymbol{\omega}^2+\gamma^2)}$.
    
    \item A Polynomial kernel $k_{\rm poly}$ does not hold the $\ell_1$ nor $\ell_2$ distance-preserving property and is defined based on the inner product as follows:
    \begin{equation}
    \label{eq:kernel-poly}
    k_{\rm poly}(\boldsymbol{z}_1,\boldsymbol{z}_2)=(\boldsymbol{z}_1^\top\boldsymbol{z}_2+c)^d.
    \end{equation}
    $k_{\rm poly}$ is not shift-invariant and thereby the RFFs cannot be employed to approximate $k_{\rm poly}$. 
    We adopt another technique to obtain an explicit random mapping for $k_{\rm poly}$, namely the Tensor Sketch approximation \cite{pham2013fast}.
\end{itemize}
We evaluate these individual kernels and their concatenations with the Cosine kernel for OoD detection; results are compared comprehensively in Table \ref{tab:exp-varied-kernels}.
All experiments are executed on the ImageNet-1K benchmark and the PyTorch-released ResNet50 model.
We can have the following observations.

\paragraph{The Cosine kernel is indispensable for OoD detection}
Without the Cosine kernel, these individual Polynomial, Laplacian, and Gaussian kernels all show a substantial detection performance drop.
Besides, KPCA w.r.t. a single Cosine kernel significantly enhances PCA.
It implies that feature normalization via the Cosine kernel is essential for learning a discriminative subspace to separate InD and OoD data.

\paragraph{The Gaussian kernel shows the best detection results on top of the Cosine kernel}
In Table \ref{tab:exp-varied-kernels}, both Cosine-Laplacian kernel and Cosine-Polynomial kernel cannot bring detection performance gains compared to the single Cosine kernel, which illustrates that the $\ell_1$-distance relation from $k_{\rm lap}$ and the inner-product information from $k_{\rm poly}$ in the $\phi_{\rm cos}(\boldsymbol{z})$-space actually hurt the separability between InD and OoD features.

\paragraph{The data-dependent Nystr\"om approximation outperforms the data-independent random features in OoD detection}
For the three kernels of Cosine-Polynomial, Cosine-Laplacian, and Cosine-Gaussian, KPCA via Nystr\"om achieves lower FPR and higher AUROC values than that via random features, which implies the superiority of data-dependent approximation over the data-independent way.

\begin{table}[t]
    \centering
    \caption{Ablation studies on the sampling scheme for Nystr\"om in $\rm KPCA_{nys}$. Average detection results are reported.}
    {\begin{tabular}{@{}l|cc@{}}
    \toprule
    \multirow{2}{*}{sampling scheme} & \multicolumn{2}{c}{\bf AVERAGE} \\
    & FPR$\downarrow$ & AUROC$\uparrow$ \\
    \midrule
     Uniform & 49.94 & 87.17 \\
     High-Energy & 58.23 & 83.79 \\
     Low-Energy & 38.15 & 91.02 \\
    \bottomrule
    \end{tabular}}
    \label{tab:exp-ablation-sampling}
\end{table}

\begin{table}[t]
    \centering
    \caption{Detection performance of KPCA applied to intermediate-layer features. Average detection results are reported.}
    {\begin{tabular}{@{}lc|cc@{}}
    \toprule
    \multirow{2}{*}{method} & \multirow{2}{*}{\makecell{feature\\position}} & \multicolumn{2}{c}{{\bf AVERAGE}} \\
    & & FPR$\downarrow$ & AUROC$\uparrow$ \\
    \midrule
    \multirow{4}{*}{$\rm KPCA_{rff}$} & \texttt{layer1} & 80.87 & 67.42 \\
    & \texttt{layer2} & 79.69 & 67.52 \\
    & \texttt{layer3} & 72.38 & 74.72 \\
    & \texttt{layer4} & 48.60 & 87.53 \\
     \midrule
     \multirow{4}{*}{$\rm KPCA_{nys}$} & \texttt{layer1} & 80.48 & 67.86 \\
     & \texttt{layer2} & 79.96 & 68.26 \\
     & \texttt{layer3} & 61.59 & 80.53 \\
     & \texttt{layer4} & 38.15 & 91.02 \\
    \bottomrule
    \end{tabular}}
    \label{tab:exp-interlayer-feat}
\end{table}

\subsection{Ablation Studies}

\noindent{\bf Effect of different sampling schemes.}\quad
In our $\rm KPCA_{nys}$, an energy-based sampling scheme is specifically devised for Nystr\"om, which selects those low-energy InD training samples $\{\boldsymbol{\tilde x}_i\}_{i=1}^{M_n}$ with the smallest-$M_n$ energy values as support landmarks.
This sampling scheme shows to produce discriminative non-linear principal components capturing InD-OoD disparities and improved detection results.
Ablations are studied in this section to verify its effectiveness.
Particularly, we consider another two sampling configurations as follows:
\begin{itemize}
    \item A {\it uniform} sampling is widely adopted in Nystr\"om \cite{NIPS2000_19de10ad,JMLR:v6:drineas05a} and is investigated in OoD detection for ablations.
    \item Regarding the proposed energy-based sampling scheme, we consider those high-energy InD training samples with the {\it largest}-$M_n$ energy values as landmarks for ablations.
\end{itemize}
Comparisons among different sampling schemes are presented in Table \ref{tab:exp-ablation-sampling}, with full results in Table \ref{response:tab:full-auroc} of Appendix \ref{sec:app:exp-supp-ood}.

As illustrated, our low-energy support landmarks outperform the uniformly-sampled landmarks for $\rm KPCA_{nys}$, which indicates that appropriate landmarks in relation to this specific OoD detection task can bring detection performance gains.
Besides, low-energy landmarks show to be superior over high-energy ones in separating InD and OoD with $\rm KPCA_{nys}$.
The reason can be that the approximation based on those boundary InD samples leads to reconstruction errors that are more sensitive to distribution shifts, as discussed in Section \ref{sec:method:nys}.

\noindent{\bf Effect of intermediate-layer features.}\quad
While our subspace learning is typically applied to penultimate-layer features, this study extends the analysis by evaluating KPCA on intermediate-layer features of a ResNet50 model to assess their discriminative capability.
A typical ResNet50 architecture\footnote{\href{https://docs.pytorch.org/vision/main/\_modules/torchvision/models/resnet.html\#resnet50}{https://docs.pytorch.org/vision/main/\_modules/torchvision/models/resnet.\-html\#resnet50}} consists of one beginning convolution block, followed by four sequential \texttt{layer} blocks (\texttt{layer1}, \texttt{layer2}, \texttt{layer3} and \texttt{layer4}), and a final linear classification layer.
Beyond its default use at \texttt{layer4}, we apply KPCA to features from earlier blocks (\texttt{layer1}, \texttt{layer2} and \texttt{layer3}) to investigate performance in their respective non-linear subspaces with results shown in Table \ref{tab:exp-interlayer-feat}.

As demonstrated, shallow-layer features (\texttt{layer1} and \texttt{layer2}) does not yield effective non-linear subspaces to well separate InD and OoD data, while deeper features (\texttt{layer3} and \texttt{layer4}) exhibit enhanced detection results.
This observation aligns with the hierarchical learning of DNNs: shallow layers capture fine-grained image patterns and higher-level semantic discrimination emerges in deeper layers.
Consequently, the penultimate-layer features (\texttt{layer4}), adopted in our work by default, demonstrate the strongest discriminative ability and achieve the best detection performance for both $\rm KPCA_{rff}$ and $\rm KPCA_{nys}$.
This choice is also supported by prior findings that penultimate-layer feature collapse can aid OoD detection \cite{van2020uncertainty}.

\subsection{Sensitivity Analysis}
\label{sec:exp:analy-sensitivity}

A comprehensive sensitivity analysis is executed to illustrate the effects of hyper-parameters in our KPCA detection framework.
$\rm KPCA_{rff}$ and $\rm KPCA_{nys}$ share two common hyper-parameters: the bandwidth $\gamma$ of the Gaussian kernel and the number of columns $q$ of the projection matrix ${\bf U}^\Phi_q$.
Besides, their distinctive hyper-parameters are the number of RFFs $M_r$ in $\rm KPCA_{rff}$ and the number of support landmarks $M_n$ in $\rm KPCA_{nys}$.
In the following, we discuss the influence of each hyper-parameter, and report detection results by varying one hyper-parameter with the others fixed.
All experiments are conducted on the ImageNet-1K benchmark and the PyTorch-released ResNet50 checkpoint.

\noindent{\bf Effect of $q$.}\quad
The column size $q$ in the projection matrix ${\bf U}^\Phi_q$ indicates the number of preserved principal components and determines the amount of information captured by the projected subspace for InD and OoD data.
Generally, $q$ is selected as the minimum number of principal components required to capture a cumulative {\it explained variance ratio} that exceeds a given threshold.
Thereby, we investigate the effect of $q$ by exploring the detection performance under varied values of the explained variance ratio, which is calculated as the ratio of the sum of eigenvalues w.r.t. the preserved principal components over the sum of all eigenvalues.

\begin{figure}[!t]
    \centering
    
    \includegraphics[width=0.85\linewidth]{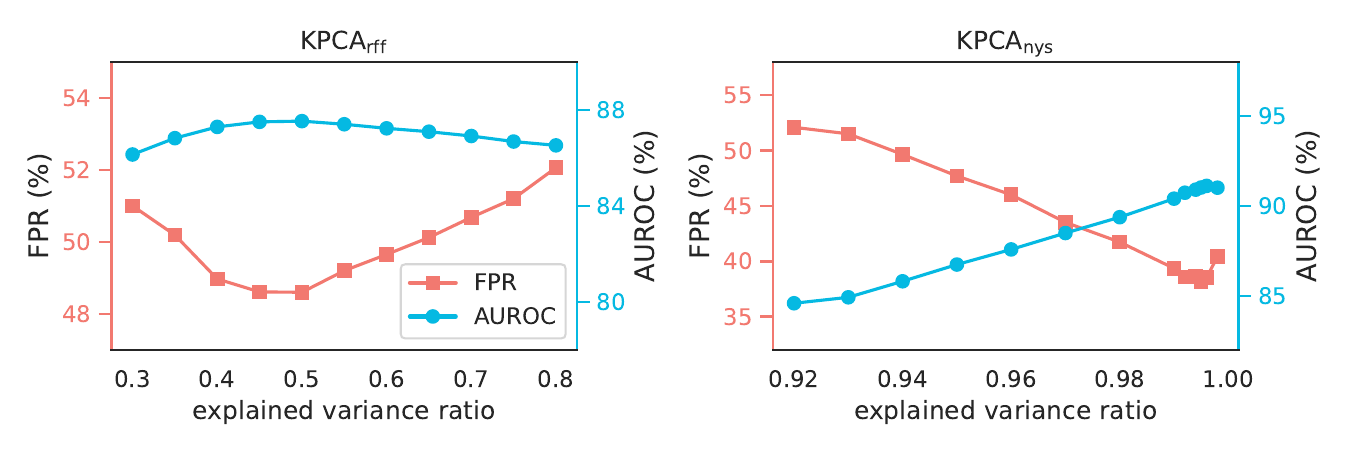}
    
    \caption{A sensitivity analysis on the explained variance ratio of $\rm KPCA_{rff}$ (left) and $\rm KPCA_{nys}$ (right). The average FPR and AUROC values of OoD datasets in the ImageNet-1K benchmark are reported. The Gaussian kernel width $\gamma$ and the numbers of RFFs and landmarks $M_r,M_n$ are fixed.}
    \label{fig:exp-sensitivity-exp-var-ratio}
\end{figure}

Figure \ref{fig:exp-sensitivity-exp-var-ratio} illustrates the detection performance of $\rm KPCA_{rff}$ and $\rm KPCA_{nys}$ under varied explained variance ratios. 
For $\rm KPCA_{nys}$, a sufficiently large value of the explained variance ratio, i.e., over 99\%, is essential to keep most principal components for superior detection results. 
Regarding $\rm KPCA_{rff}$, a modest value of the explained variance ratio is instead suggested, which might be due to the fact that the random features make the useful information for distinguishing OoD samples concentrated in fewer principal components.

\begin{figure}[t]
\centering
\includegraphics[width=0.85\linewidth]{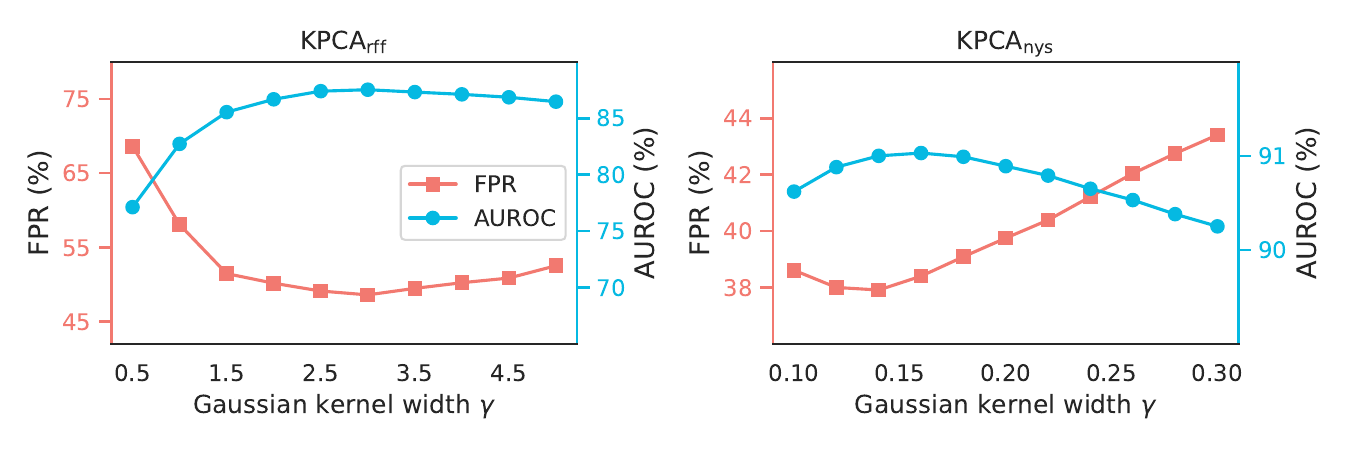}
\caption{A sensitivity analysis on the Gaussian kernel width $\gamma$ of $\rm KPCA_{rff}$ (left) and $\rm KPCA_{nys}$ (right). The average FPR and AUROC values of OoD datasets in the ImageNet-1K benchmark are reported. 
The explained variance ratio and the numbers of RFFs and landmarks $M_r,M_n$ are fixed.
}
\label{fig:exp-sensitivity-gm}
\end{figure}

\begin{figure}[t]
\centering
\includegraphics[width=0.85\linewidth]{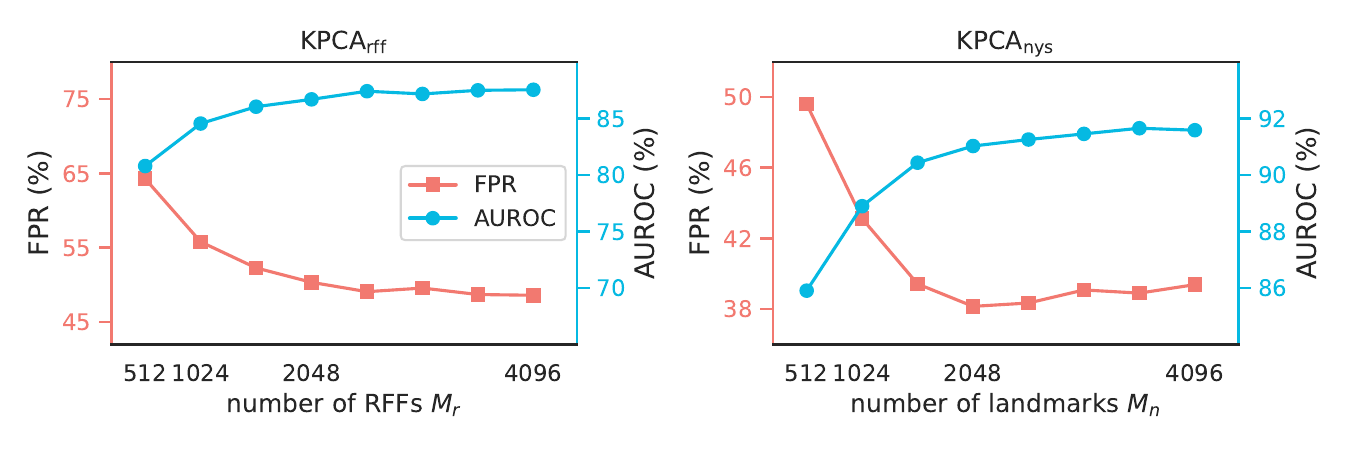}
\caption{A sensitivity analysis on the number of RFFs $M_r$ of $\rm KPCA_{rff}$ (left) and the number of landmarks $M_n$ of $\rm KPCA_{nys}$ (right). The average results of OoD datasets in the ImageNet-1K benchmark are reported. The explained variance ratio and the Gaussian kernel width $\gamma$ are fixed.
}
\label{fig:exp-sensitivity-M}
\end{figure}

\noindent{\bf Effect of $\gamma$.}\quad 
The Gaussian kernel width $\gamma$ directly affects the mapped data distribution.
For a large $\gamma$, $k_{\rm gau}(\boldsymbol{z}_1,\boldsymbol{z}_2)=e^{-\gamma\left\Vert\boldsymbol{z}_1-\boldsymbol{z}_2\right\Vert_2^2}\approx0$ for $\boldsymbol{z}_1\neq\boldsymbol{z}_2$, which indicates that the mappings of $\boldsymbol{z}_1$ and $\boldsymbol{z}_2$ are (nearly) mutually-orthogonal.
In this case, a PCA in the mapped space would become meaningless.
For a small $\gamma$, KPCA reconstruction errors will approach standard PCA ones, as discussed in \cite{hoffmann2007kernel}.

Figure \ref{fig:exp-sensitivity-gm} demonstrates the FPR and AUROC values of varied Gaussian kernel widths $\gamma$ in $\rm KPCA_{rff}$ and $\rm KPCA_{nys}$.
Clearly, neither a too large nor a too small kernel width can well characterize the patterns of InD-OoD disparities, and a mild value of $\gamma$ should be carefully tuned in both $\rm KPCA_{rff}$ and $\rm KPCA_{nys}$ for outstanding detection performance.
Moreover, it can be observed that $\rm KPCA_{nys}$ is significantly more sensitive to variations in $\gamma$ compared to $\rm KPCA_{rff}$.
This difference may stem from the fact that the low-rank approximation based on selected InD landmark InD requires more precise kernel configurations to effectively learn discriminative subspaces than the data-independent random features approximation.

\noindent{\bf Effect of $M_r,M_n$.}\quad
In $\rm KPCA_{rff}$ and $\rm KPCA_{nys}$, the number of RFFs $M_r$ and the number of support landmarks $M_n$ respectively determine the approximation ability of random features to the kernel function and of the low-rank matrix to the full kernel matrix.
Thereby, the larger the $M_r$ and $M_n$, the better random features and low-rank features approximate the Gaussian kernel $k_{\rm gau}$ \cite{rahimi2007random,NIPS2000_19de10ad}.

Figure \ref{fig:exp-sensitivity-M} indicates the FPR and AUROC values of $\rm KPCA_{rff}$ and $\rm KPCA_{nys}$ under varied numbers of RFFs and support landmarks $M_r,M_n$.
As $M_r$ and $M_n$ increase, the detection performance of both $\rm KPCA_{rff}$ and $\rm KPCA_{nys}$ improves, since the random features and low-rank features gradually converge to the Gaussian kernel.
Besides, Figure \ref{fig:exp-sensitivity-M} also provides supplementary results to discussions on the computational complexity in Section \ref{sec:exp:comput-complex}.
Under the same value of $M_r$ and $M_n$, the low-rank mapping of $\rm KPCA_{nys}$ significantly outperforms the random mapping of $\rm KPCA_{rff}$ in detection performance.
We can further find in Figure \ref{fig:exp-sensitivity-M} that the low-rank mapping built by sampling 1,024 support landmarks even exceeds the detection performance of the 4,096-dimensional random mapping, which strongly supports the superiority of the data-dependent low-rank approximation in OoD detection.

\section{Related Work}
\label{sec:related-work}

Prevalent OoD detection methods devise the scoring function $S(\cdot)$ of Eqn.\eqref{eq:ood-scoring} based on different outputs from DNNs to characterize the InD-OoD disparities, categorized as follows.

\textbf{Logits-based} approaches exploit the abnormal responses reflected in the predictive logits or probabilities from DNNs to detect OoD data.
Early methods adopt either the maximum logits \cite{hendrycks2022scaling} or probability \cite{hendrycks2016baseline,liang2018enhancing} or the energy function on logits \cite{liu2020energy} as the detection score.
Besides, decoupling on logits is explored to regularize the model training from the perspectives of InD confidence \cite{hsu2020generalized} and similarity \cite{zhang2023decoupling}, respectively.
The Wasserstein distance \cite{wang2023wood} is applied to the predictive probabilities and leads to a novel loss function to optimize the network for enhanced detection performance.

\textbf{Gradients-based} methods investigate gradient disparities between InD and OoD data for OoD detection.
Norms of full gradients are firstly studied in \cite{huang2021importance}.
The projections of gradients onto a low-rank feature space are investigated in \cite{behpour2023gradorth} for more discriminative gradient norms.
Gradient information is also utilized to regularize the model training together with auxiliary sampled OoD data \cite{sharifi2024gradient}.
A recent work \cite{wu2024low} leverages low-dimensional gradient representations to train an auxiliary linear network for effective detection.

\textbf{Features-based} approaches capture the feature information causing over-confident OoD predictions in different ways.
A series of feature rectification methods clip the abnormal feature values \cite{sun2021react,zhu2022boosting,xu2023vra,djurisicextremely,xuscaling,yuan2024discriminability,ahn2023line,song2022rankfeat} to remove anomalous information in distinct ways.
Besides, feature distances in different forms, such as the Mahalanobis distance \cite{lee2018simple,mullermahalanobis++}, the $\ell_2$ distance \cite{sun2022out,park2023nearest}, the tangent distance \cite{li2024characterizing}, and the Hopfield energy \cite{zhang2022out}, can be efficacious detection metrics.
More explorations on feature norms \cite{wang2022vim,yu2023block} and feature subspaces \cite{guan2023revisit,chen2023wdiscood,fang2024kpca,zongur2025activation}, have exhibited excellent detection performance.

Aside from the aforementioned 3 types of detection methods, a broad variety of researches have been developed for investigating InD-OoD disparities from distinctive perspectives with novel insights, including but not limited to utilizing the parameters \cite{sun2022dice,song2024rankfeat}, the ensemble technique \cite{fang2024revisiting}, the diffusion models \cite{liu2023unsupervised}, and the contrastive learning \cite{tack2020csi}.
In addition to those practical approaches, theoretical understandings for OoD detection are discussed in \cite{fang2022out,morteza2022provable}.
OoD detection is also extended in more scenarios, such as applications on the tabular data \cite{zhao2025out} and balancing the InD generalization \cite{zhao2023supervision}.

\textbf{Random Fourier Features} (RFFs) \cite{rahimi2007random}, as explicit mappings to approximate kernels, are initially developed to accelerate large-scale kernel machines, and have been extensively studied and applied across a broad range of fields, such as kernel approximation \cite{liu2021random}, kernel learning \cite{fang2023end}, generalization \cite{yang2012nystrom}, optimization \cite{belkin2019reconciling} and deep network architectures \cite{li2022convolutional}.

\textbf{Nystr\"om} method \cite{NIPS2000_19de10ad} with various sampling schemes \cite{kumar2012sampling} has been applied not only to kernels, such as kernel learning \cite{JMLR:v6:drineas05a,tao2024learning} and approximated KPCA \cite{lopez2014randomized,ullah2018streaming,sterge2020gain,hallgren2021kernel,sriperumbudur2022approximate,sterge2022statistical}, but also to diverse domains beyond kernels, e.g., interpreting the self-attention in transformers \cite{xiong2021nystromformer} and differentiation optimization \cite{hataya2023nystrom}.
There is significant potential in bridging these classic machine learning methods with modern deep learning frameworks, leveraging their theoretical foundations and computational efficiency to address contemporary challenges.

\section{Conclusion and Discussion}
\label{sec:conclusion}

Exploring the non-linear feature subspace remains a fresh, significant, and open perspective for OoD detection.
In this work, we leverage the framework of KPCA to identify discriminative non-linear feature subspaces to detect OoD data in an efficacious and efficient way.
Firstly, we analyze InD and OoD features, observe the non-linear patterns related to InD-OoD disparities, and derive an effective Cosine-Gaussian kernel that models such patterns.
Then, two distinct explicit mappings $\Phi(\cdot)$ to approximate the Cosine-Gaussian kernel are leveraged: data-independent random features and data-driven Nystr\"om approximation, so as to achieve efficient computations of the KPCA reconstruction errors in the $\Phi(\boldsymbol{z})$-space.
Particularly, an energy-based sampling scheme is proposed to refine the Nystr\"om method for more discriminative subspaces between InD and OoD.
Extensive experiments verify that our KPCA method achieves new SOTA OoD detection performance with high efficiency, and provide in-depth analyses on kernel choices, sampling schemes, intermediate-layer features, and hyper-parameters.

One limitation of KPCA detection is the hyper-parameter tuning of its task-specific Cosine-Gaussian kernel for non-linear mappings.
Our preliminary explorations on learnable kernel representations in Section \ref{sec:method:kernel-learning} and Appendix \ref{sec:app:learnable-kpca} mitigate this issue through different learning techniques, while also suggesting further improvements, such as incorporating outliers and tailored optimization objectives for distribution shifts.
We hope that the proposed effective kernel verified empirically in our work could benefit the research community from the perspective of exploring the non-linearity in feature space in future studies on OoD detection.

\bibliographystyle{unsrt}  
\bibliography{reference}

\clearpage

\setcounter{table}{0}
\renewcommand*{\thetable}{S\arabic{table}}
\setcounter{equation}{0}
\renewcommand*{\theequation}{S\arabic{equation}}

\begin{appendices}

\section{Proof}
\label{sec:app:proof}
The proof of Proposition \ref{thm:reconstruction-error-kpca} is presented.
\begin{proof}
Recall $\boldsymbol{z}\in\mathbb{R}^m$ and suppose the feature mapping {\it w.r.t} the kernel $k$ is denoted as $\Phi:\mathbb{R}^m\rightarrow\mathbb{R}^M$ with $k(\boldsymbol{z}_1,\boldsymbol{z}_2)=\Phi(\boldsymbol{z}_1)^\top\Phi(\boldsymbol{z}_2)$.
Given a finite training set $\{\boldsymbol{z}_i\}_{i=1}^{N_{\rm tr}}$, eigendecomposition is executed on the covariance matrix on the mapped training set $\{\Phi(\boldsymbol{z}_i)\}_{i=1}^{N_{\rm tr}}$, leading to the left eigenmatrix ${\bf U}^\Phi\in\mathbb{R}^{M\times M}$.
By taking the first $q$ columns in ${\bf U}^\Phi$, we obtain the projection matrix ${\bf U}^\Phi_q$ with ${\bf U}^\Phi=\left[{\bf U}^\Phi_q,{\bf U}^\Phi_p\right]$.

For the reconstruction error $e^\Phi(\boldsymbol{\hat z})$ of a new test sample $\boldsymbol{\hat z}\in\mathbb{R}^m$ in the mapped $\Phi(\boldsymbol{z})$-space, we have:

\begin{equation}
\begin{aligned}
\label{eq:reconstruction-error-residual-proof}
\left(e^{\Phi}(\boldsymbol{\hat z})\right)^2
&=\left\Vert
(\Phi(\boldsymbol{\hat z})-\boldsymbol{\mu}^\Phi)-{\bf U}^\Phi_q{\bf U}_q^{\Phi\top}(\Phi(\boldsymbol{\hat z})-\boldsymbol{\mu}^\Phi)
\right\Vert_2^2\\
&=\left((\Phi(\boldsymbol{\hat z})-\boldsymbol{\mu}^\Phi)-{\bf U}^\Phi_q{\bf U}_q^{\Phi\top}(\Phi(\boldsymbol{\hat z})-\boldsymbol{\mu}^\Phi)\right)^\top
\left((\Phi(\boldsymbol{\hat z})-\boldsymbol{\mu}^\Phi)-{\bf U}^\Phi_q{\bf U}_q^{\Phi\top}(\Phi(\boldsymbol{\hat z})-\boldsymbol{\mu}^\Phi)\right)\\
&=(\Phi(\boldsymbol{\hat z})-\boldsymbol{\mu}^\Phi)^\top(\Phi(\boldsymbol{\hat z})-\boldsymbol{\mu}^\Phi)
-(\Phi(\boldsymbol{\hat z})-\boldsymbol{\mu}^\Phi)^\top{\bf U}^\Phi_q{\bf U}_q^{\Phi\top}(\Phi(\boldsymbol{\hat z})-\boldsymbol{\mu}^\Phi)\\
&=(\Phi(\boldsymbol{\hat z})-\boldsymbol{\mu}^\Phi)^\top
\left({\bf I}-{\bf U}^\Phi_q{\bf U}_q^{\Phi\top}\right)
(\Phi(\boldsymbol{\hat z})-\boldsymbol{\mu}^\Phi)\\
&=(\Phi(\boldsymbol{\hat z})-\boldsymbol{\mu}^\Phi)^\top
{\bf U}^\Phi_p{\bf U}_p^{\Phi\top}
(\Phi(\boldsymbol{\hat z})-\boldsymbol{\mu}^\Phi)\\
&=\left\Vert{\bf U}_p^{\Phi\top}(\Phi(\boldsymbol{\hat z})-\boldsymbol{\mu}^\Phi)
\right\Vert_2^2.
\end{aligned}
\end{equation}
We have $e^\Phi(\boldsymbol{\hat z})=\left\Vert{\bf U}_p^{\Phi\top}(\Phi(\boldsymbol{\hat z})-\boldsymbol{\mu}^\Phi)\right\Vert_2$.
In this sense, the reconstruction error is interpreted as the norm of features projected into the residual subspace, i.e., the ${\bf U}_p$-space.

Therefore, to obtain KPCA reconstruction errors via the kernel matrix $\bf K$, we can employ the kernel trick to project $\boldsymbol{\hat z}$ into the residual space through $\bf K$, as typically conducted for dimension reduction through KPCA \cite{scholkopf1997kernel,scholkopf1998nonlinear}.
Accordingly, the KPCA reconstruction error $e^k$ is calculated as the norm of features projected into the residual space w.r.t. the eigenvectors of small eigenvalues of the kernel matrix $\bf K$, as stated in Proposition \ref{thm:reconstruction-error-kpca}. The proof is completed.
\end{proof}

\section{Additional Experiments on OoD detection}
\label{sec:app:exp-supp-ood}

In this section, additional experiments on OoD detection are provided, including results of MobileNetV2 \cite{sandler2018mobilenetv2} on the ImageNet-1K \cite{deng2009imagenet} benchmark, results of ResNet18 \cite{he2016deep} on the CIFAR10 \cite{krizhevsky2009learning} benchmark, and full results on each OoD dataset w.r.t. Tables \ref{tab:exp-imgnet-r50-fuse}, \ref{tab:exp-varied-kernels}, \ref{tab:exp-ablation-sampling} in the main text.

\subsection{Experiments on MobileNet}
\label{sec:exp-ood-ood-imgnet}
Table \ref{tab:exp-imgnet-mnet-fuse} supplements detection results on the ImageNet-1K benchmark and MobileNetV2.
We use the pre-trained checkpoint of MobileNetV2 released by PyTorch \cite{paszke2019pytorch}.
The KPCA detection method is enhanced via the pioneering feature rectification baseline ReAct \cite{sun2021react}. 

Similar to the results of ResNet50 in Table \ref{tab:exp-imgnet-r50-fuse} of the main text, $\rm KPCA_{rff}$ and $\rm KPC_{nys}$ equipped with ReAct outperform a wide variety of different detection methods that are either feature rectification or combined with feature rectification, indicating the superiority of exploring the non-linearity in the feature space via suitable kernels on different network structures.

\begin{table*}[ht]
    \centering
    \caption{Comparisons for KPCA enhanced by feature pruning. Results are of {\bf MobileNetV2} on {\bf ImageNet-1K}.}
    \resizebox{\textwidth}{!}{
    \begin{tabular}{l|cc cc cc cc|cc}
    \toprule
    \multirow{3}{*}{method} & \multicolumn{8}{c|}{OoD data sets} & \multicolumn{2}{c}{\multirow{2}{*}{\bf AVERAGE}} \\
    & \multicolumn{2}{c}{iNaturalist} & \multicolumn{2}{c}{SUN} & \multicolumn{2}{c}{Places} & \multicolumn{2}{c|}{Textures} & & \\
    & FPR$\downarrow$ & AUROC$\uparrow$ & FPR$\downarrow$ & AUROC$\uparrow$ &
    FPR$\downarrow$ & AUROC$\uparrow$ & FPR$\downarrow$ & AUROC$\uparrow$ & FPR$\downarrow$ & AUROC$\uparrow$ \\
    \midrule
    MSP \cite{hendrycks2016baseline} & 64.29 & 85.32 & 77.02 & 77.10 & 79.23 & 76.27 & 73.51 & 77.30 & 73.51 & 79.00 \\
    Energy \cite{liu2020energy} & 59.50 & 88.91 & 62.65 & 84.50 & 69.37 & 81.19 & 58.05 & 85.03 & 62.39 & 84.91 \\
    ODIN \cite{liang2018enhancing} & 58.54 & 87.51 & 57.00 & 85.83 & 59.87 & 84.77 & 52.07 & 85.04 & 56.87 & 85.79 \\
    Mahala \cite{lee2018simple} & 62.11 & 81.00 & 47.82 & 83.66 & 52.09 & 83.63 & 92.38 & 33.06 & 63.60 & 71.01 \\
    ViM \cite{wang2022vim} & 91.83 & 77.47 & 94.34 & 70.24 & 93.97 & 68.26 & 37.62 & 92.65 & 79.44 & 77.15 \\
    DICE \cite{sun2022dice} & 43.28 & 90.79 & 38.86 & 90.41 & 53.48 & 85.67 & 33.14 & 91.26 & 42.19 & 89.53 \\
    ReAct \cite{sun2021react} & 43.07 & 92.72 & 52.47 & 87.26 & 59.91 & 84.07 & 40.20 & 90.96 & 48.91 & 88.75 \\
    BATS \cite{zhu2022boosting} & 49.57 & 91.50 & 57.81 & 85.96 & 64.48 & 82.83 & 39.77 & 91.17 & 52.91 & 87.87 \\
    ASH-B \cite{djurisicextremely} & 31.46 & 94.28 & 38.45 & 91.61 & 51.80 & 87.56 & 20.92 & 95.07 & 35.66 & 92.13 \\
    NNGuide \cite{park2023nearest} & 45.73 & 91.19 & 51.03 & 87.87 & 60.60 & 84.44 & 29.50 & 92.47 & 46.72 & 88.99 \\
    GradOrth \cite{behpour2023gradorth} & 26.81 & 93.17 & 30.82 & 93.18 & 40.27 & 89.12 & 12.69 & {\bf97.52} & 27.65 & 93.25 \\
    DICE \cite{sun2022dice}+ReAct & 41.75 & 89.84 & 39.07 & 90.39 & 54.41 & 84.03 & 19.98 & 95.86 & 38.80 & 90.03 \\
    LowDim \cite{wu2024low}+ReAct & 34.80 & 93.22 & 50.09 & 89.44 & 57.05 & 87.02 & 40.19 & 90.91 & 45.53 & 90.15 \\
    NNGuide \cite{park2023nearest}+ReAct & 37.87 & 92.83 & 43.51 & 89.57 & 55.92 & 85.24 & 14.34 & 96.30 & 37.91 & 90.99 \\
    RegPCA \cite{guan2023revisit}+ReAct & 35.84 & 93.66 & 40.35 & 90.77 & 52.38 & 86.76 & 18.44 & 95.39 & 36.75 & 91.65 \\
    \rowcolor{tabgray}{\bf KPCA}$_{\rm rff}$+ReAct & {31.72} & {94.27} & 40.77 & {90.98} & 55.69 & 86.42 & {\bf10.48} & {97.49} & {34.66} & {92.29} \\
    \rowcolor{tabgray}{\bf KPCA}$_{\rm nys}$+ReAct & {\bf23.69} & {\bf95.83} & {\bf22.73} & {\bf95.55} & {\bf38.19} & {\bf91.91} & 21.08 & 95.87 & {\bf26.42} & {\bf94.79} \\
    \bottomrule
    \end{tabular}}
    \label{tab:exp-imgnet-mnet-fuse}
\end{table*}

\begin{table}[t]
    \centering
    \caption{Comparisons among a wide variety of OoD detection methods with \textbf{ResNet18} models trained on \textbf{CIFAR10}. The FPR and AUROC values are averaged over the selected multiple OoD datasets.}
    \begin{tabular}{ll|cc}
    \toprule
    training & method & FPR$\downarrow$ & AUROC$\uparrow$ \\
    \midrule
    \multirow{8}{*}{\makecell[l]{Standard\\Training}} & MSP \cite{hendrycks2016baseline} & 57.67 & 90.86 \\
    &ODIN \cite{liang2018enhancing} & 36.16 & 92.30 \\
    &Energy \cite{liu2020energy} & 38.02 & 93.05 \\
    &GODIN \cite{hsu2020generalized} & 32.80 & 93.52 \\
    &Mahala \cite{lee2018simple} & 37.94 & 86.50 \\
    &KNN \cite{sun2022out} & 30.77 & 94.15 \\
    &\cellcolor{tabgray}{\bf KPCA}$_{\rm rff}$ & \cellcolor{tabgray}{27.34} & \cellcolor{tabgray}{94.95} \\
    &\cellcolor{tabgray}{\bf KPCA}$_{\rm nys}$ & \cellcolor{tabgray}{\bf25.73} & \cellcolor{tabgray}{\bf95.25} \\
    \midrule
    \multirow{6}{*}{\makecell[l]{Regularized\\Training}} & CSI \cite{tack2020csi} & 24.16 & 95.89 \\
    &SSD+ \cite{sehwag2020ssd} & 16.52 & 97.15 \\
    &KNN+ \cite{sun2022out} & 11.07 & 97.93 \\
    &TNN \cite{li2024characterizing} & 9.87 & 98.05 \\
    &\cellcolor{tabgray}{\bf KPCA}$_{\rm rff}$ & \cellcolor{tabgray}{\bf 7.35} & \cellcolor{tabgray}{\bf 98.37} \\
    &\cellcolor{tabgray}{\bf KPCA}$_{\rm nys}$ & \cellcolor{tabgray}7.64 & \cellcolor{tabgray}98.31 \\
    \bottomrule
    \end{tabular}
    \label{tab:exp-c10-r18}
\end{table}

\subsection{Experiments on CIFAR10}
\label{sec:exp-supp-ood-c10}

Extensive experiments on the more realistic and challenging large-scale ImageNet-1K benchmark have demonstrated the advantageous detection performance and computational efficiency of our KPCA method.
In addition, we further supplement results on the small-scale CIFAR10 benchmark for reference.
For CIFAR10 as InD, OoD datasets include SVHN \cite{netzer2011reading}, LSUN \cite{yu2015lsun}, iSUN \cite{xu2015turkergaze}, Textures \cite{cimpoi2014describing} and Places365 \cite{zhou2017places}.
Following the settings in KNN \cite{sun2022out}, we choose ResNet18 models \cite{he2016deep} that are trained via standard cross-entropy loss and supervised contrastive learning \cite{khosla2020supervised}, and adopt the checkpoints released by KNN for a fair comparison.
Images from CIFAR10 and the paired OoD datasets are resized to $32\times32\times3$ as inputs for ResNet18.

As the kernel is inspired by the KNN approach \cite{sun2022out}, our KPCA detection method is primarily compared with KNN, and the comparisons follow the reported configurations in KNN \cite{sun2022out}, shown in Table \ref{tab:exp-c10-r18}.
For both standard and regularized training, $\rm KPCA_{rff}$ and $\rm KPCA_{nys}$ outperform KNN, highlighting the advantage of non-linear kernels in capturing the disparities between InD and OoD data in the feature space.
Particularly, under regularized training, $\rm KPCA_{rff}$ and $\rm KPCA_{nys}$ exhibit nearly the same detection performance, which demonstrates that the random features and low-rank features achieve comparable approximation quality given features derived from supervised contrastive learning on such small-scale data.
The superiority of low-rank features is mainly pronounced on more complex datasets, as illustrated by results on the ImageNet-1K benchmark in the main text.

\subsection{Full Results}

We provide the full detection results for each OoD dataset, corresponding to the three main experiments in the main text.
\begin{itemize}
    \item Table \ref{tab:exp-imgnet-r50-fuse-full} presents the extended results of KPCA with feature rectification, complementing Table \ref{tab:exp-imgnet-r50-fuse}.
    \item Table \ref{tab:exp-varied-kernels-full} gives a complete ablation study on kernel designs, extending Table \ref{tab:exp-varied-kernels}.
    \item Table \ref{response:tab:full-auroc} lists the detailed results for different Nystr\"om sampling strategies, supplementing Table \ref{tab:exp-ablation-sampling}.
\end{itemize}

\begin{table*}[t]
    \centering
    \caption{Detection performance of KPCA enhanced by feature rectification. Results are of {\bf ResNet50} on {\bf ImageNet-1K}.}
    \resizebox{\textwidth}{!}{
    \begin{tabular}{l|cc cc cc cc|cc}
    \toprule
    \multirow{3}{*}{method} & \multicolumn{8}{c|}{OoD data sets} & \multicolumn{2}{c}{\multirow{2}{*}{\bf AVERAGE}} \\
    & \multicolumn{2}{c}{iNaturalist} & \multicolumn{2}{c}{SUN} & \multicolumn{2}{c}{Places} & \multicolumn{2}{c|}{Textures} & & \\
    & FPR$\downarrow$ & AUROC$\uparrow$ & FPR$\downarrow$ & AUROC$\uparrow$ &
    FPR$\downarrow$ & AUROC$\uparrow$ & FPR$\downarrow$ & AUROC$\uparrow$ & FPR$\downarrow$ & AUROC$\uparrow$ \\
    \midrule
    DICE \cite{sun2022dice} & 26.66 & 94.49 & 36.08 & 90.98 & 47.63 & 87.73 & 32.46 & 90.46 & 35.71 & 90.92 \\
    ReAct \cite{sun2021react} & 20.38 & 96.22 & 24.20 & 94.20 & 33.85 & 91.58 & 47.30 & 89.80 & 31.43 & 92.95\\
    BATS \cite{zhu2022boosting} & 12.57 & 97.67 & 22.62 & 95.33 & 34.34 & 91.83 & 38.90 & 92.27 & 27.11 & 94.28 \\
    VRA \cite{xu2023vra} & 15.70 & 97.12 & 26.94 & 94.25 & 37.85 & 91.27 & 21.47 & 95.62 & 25.49 & 94.57 \\
    ASH-B \cite{djurisicextremely} & 14.21 & 97.32 & 22.08 & 95.10 & 33.45 & 92.31 & 21.17 & 95.50 & 22.73 & 95.06 \\
    SCALE \cite{xuscaling} & 9.50 & 98.17 & 23.27 & 95.02 & 34.51 & 92.26 & 12.93 & {97.37} & 20.05 & 95.71 \\
    LINe \cite{ahn2023line} & 22.52 & 94.44 & 19.48 & 95.26 & 12.24 & 97.56 & 28.54 & 92.84 & 20.69 & 95.03 \\
    DDCS \cite{yuan2024discriminability} & 11.63 & 97.85 & 18.63 & 95.68 & 28.78 & 92.89 & 18.40 & 95.77 & 19.36 & 95.55 \\
    GradOrth\cite{behpour2023gradorth} & 11.04 & 98.00 & 19.61 & 95.76 & 33.67 & 91.78 & {\bf11.19} & {\bf98.06} & 18.57 & 96.31 \\
    DICE \cite{sun2022dice}+ReAct & 20.08 & 96.11 & 26.50 & 93.83 & 38.34 & 90.61 & 29.36 & 92.65 & 28.57 & 93.30 \\
    GradNorm\cite{huang2021importance}+ReAct & 14.88 & 97.01 & 25.54 & 94.19 & 36.49 & 91.12 & 23.60 & 94.57 & 25.13 & 94.22 \\
    LowDim\cite{wu2024low}+ReAct & 19.87 & 95.77 & 45.34 & 90.52 & {\bf0.00} & {\bf100.00} & 26.90 & 95.50 & 23.03 & 95.45 \\
    NNGuide \cite{park2023nearest}+ReAct & 11.12 & 97.70 & 20.51 & 95.26 & 29.99 & 92.70 & 17.27 & 96.11 & 19.72 & 95.45 \\
    RegPCA\cite{guan2023revisit}+ReAct & 10.17 & 97.97 & 18.50 & 95.80 & 27.31 & 93.39 & 18.67 & 95.95 & 18.66 & 95.76 \\
    \rowcolor{tabgray}{\bf KPCA}$_{\rm rff}$+ReAct & 10.77 & 97.85 & 18.70 & 95.75 & 28.69 & 93.13 & {12.57} & 97.21 & 17.68 & 95.98 \\
    \rowcolor{tabgray}{\bf KPCA}$_{\rm nys}$+ReAct & {\bf8.83} & {\bf98.20} & {\bf12.20} & {\bf97.31} & {22.98} & {94.83} & 17.02 & 96.80 & {\bf15.26} & {\bf96.79} \\
    \bottomrule
    \end{tabular}}
    \label{tab:exp-imgnet-r50-fuse-full}
\end{table*}

\begin{table*}[t]
    \centering
    \caption{Detection results among a variety of kernels approximated via random features and the Nystr\"om approximation.}
    \resizebox{\textwidth}{!}{
    \begin{tabular}{ll|cc cc cc cc|cc}
    \toprule
    \multirow{3}{*}{kernel} & \multirow{3}{*}{approx.} & \multicolumn{8}{c|}{OoD datasets} & \multicolumn{2}{c}{\multirow{2}{*}{\bf AVERAGE}} \\
    & & \multicolumn{2}{c}{iNaturalist} & \multicolumn{2}{c}{SUN} & \multicolumn{2}{c}{Places} & \multicolumn{2}{c|}{Textures} & & \\
    & & FPR$\downarrow$ & AUROC$\uparrow$ & FPR$\downarrow$ & AUROC$\uparrow$ &
    FPR$\downarrow$ & AUROC$\uparrow$ & FPR$\downarrow$ & AUROC$\uparrow$ & FPR$\downarrow$ & AUROC$\uparrow$ \\
    \midrule
    PCA & - & 95.46 & 52.01 & 97.98 & 44.86 & 97.99 & 45.19 & 46.22 & 87.77 & 84.41 & 57.46 \\
    \midrule
    Cosine & - & 67.25 & 83.41 & 75.53 & 79.93 & 82.48 & 73.83 & 8.33 & 98.29 & 58.40 & 83.86 \\
    \multirow{2}{*}{Polynomial} & random & 96.03 & 53.07 & 98.26 & 42.84 & 97.85 & 45.02 & 95.50 & 47.96 & 96.91 & 47.22 \\
    & Nystr\"om & 92.89 & 62.45 & 96.21 & 56.71 & 97.47 & 52.75 & 58.74 & 86.68 & 86.33 & 64.65 \\
    \multirow{2}{*}{Laplacian} & random & 94.65 & 50.25 & 94.68 & 50.29 & 95.28 & 49.80 & 94.66 & 50.34 & 94.82 & 50.17 \\
    & Nystr\"om & 83.57 & 34.05 & 75.93 & 39.76 & 84.57 & 35.02 & 81.58 & 46.07 & 81.41 & 38.72 \\
    \multirow{2}{*}{Gaussian} & random & 94.46 & 50.83 & 95.17 & 50.33 & 94.80 & 50.46 & 95.09 & 50.80 & 94.88 & 50.60 \\
    & Nystr\"om & 34.11 & 37.71 & 51.44 & 28.39 & 42.22 & 33.33 & 24.82 & 42.32 & 38.15 & 35.44 \\
    \midrule
    \multirow{2}{*}{Cosine-Poly.} & random & 54.10 & 84.48 & 75.97 & 75.04 & 82.82 & 69.01 & 59.15 & 83.27 & 68.01 & 77.95 \\
    & Nystr\"om & 67.28 & 83.41 & 75.78 & 79.79 & 82.51 & 73.65 & 8.10 & 98.30 & 58.42 & 83.79 \\
    \multirow{2}{*}{Cosine-Lap.} & random & 76.18 & 77.95 & 77.54 & 76.70 & 84.47 & 70.16 & 11.97 & 97.57 & 62.54 & 80.60 \\
    & Nystr\"om & 58.49 & 84.10 & 55.72 & 87.61 & 65.61 & 82.57 & 36.37 & 92.58 & 54.05 & 86.71 \\
    \multirow{2}{*}{Cosine-Gau.} & random & 50.07 & 89.32 & 62.56 & 83.74 & 72.76 & 78.91 & 9.02 & 98.14 & 48.60 & 87.53 \\ 
    & Nystr\"om & 38.61 & 91.18 & 43.47 & 90.20 & 57.35 & 85.30 & 13.16 & 97.41 & 38.15 & 91.02 \\
    \bottomrule
    \end{tabular}}
    \label{tab:exp-varied-kernels-full}
\end{table*}

\begin{table*}[t]
    \centering
    \caption{Ablation studies on the sampling scheme for Nystr\"om in $\rm KPCA_{nys}$. Both FPR and AUROC results of each OoD dataset are listed under the 3 sampling scheme of ResNet50 trained on ImageNet-1K.}
    {\resizebox{0.9\textwidth}{!}{
    \begin{tabular}{@{}l|cc cc cc cc|cc@{}}
    \toprule
    \multirow{3}{*}{\makecell[l]{Sampling\\scheme}} & \multicolumn{8}{c|}{OoD datasets} & \multicolumn{2}{c}{\multirow{2}{*}{\bf AVERAGE}} \\
    & \multicolumn{2}{c}{iNaturalist} & \multicolumn{2}{c}{SUN} & \multicolumn{2}{c}{Places} & \multicolumn{2}{c|}{Textures} & & \\
    & FPR$\downarrow$ & AUROC$\uparrow$ & FPR$\downarrow$ & AUROC$\uparrow$ &
    FPR$\downarrow$ & AUROC$\uparrow$ & FPR$\downarrow$ & AUROC$\uparrow$ & FPR$\downarrow$ & AUROC$\uparrow$ \\
    \midrule
     Uniform & 52.28 & 88.06 & 61.50 & 84.59 & 73.59 & 78.63 & 12.38 & 97.41 & 49.94 & 87.17 \\
     High-Energy & 67.04 & 83.49 & 75.47 & 79.73 & 82.43 & 73.62 & 7.98 & 98.33 & 58.23 & 83.79 \\
     Low-Energy & 38.61 & 91.18 & 43.47 & 90.20 & 57.35 & 85.30 & 13.16 & 97.41 & 38.15 & 91.02 \\
    \bottomrule
    \end{tabular}}}
    \label{response:tab:full-auroc}
\end{table*}

\section{KPCA with Learnable Representations for OoD Detection}
\label{sec:app:learnable-kpca}

While the non-parametric Cosine-Gaussian kernel is validated with effective feature subspaces promoting InD and OoD separability, it would be interesting to further investigate the utility of learnable representations for optimizing non-linear feature subspaces.
In this section, we explore the potentials of learnable kernel representations within our KPCA-based OoD detection framework.

We note that deploying learnable representations in kernel-based methods encompasses distinct and diverse directions, e.g., using learnable kernels or parametric neural networks. We implement \textit{(i)} a Shallow AutoEncoder (Shallow-AE) network \cite{meissen2022unsupervised}, and \textit{(ii)} several representative kernel learning methods including RFF-based Optimization (RFF-OPT) \cite{sinha2016learning} and Deep Kernel Gaussian Process (DKGP) \cite{wilson2016deep} for evaluations.
The explanations for each method and the results are outlined below, respectively.

\noindent{\bf Shallow-AE} \cite{meissen2022unsupervised}. 
We consider the shallow autoencoders as mappings \cite{meissen2022unsupervised}.
Following approaches similar to \cite{meissen2022unsupervised} and for a fair comparison, an encoder $\phi_{e}$ and a decoder $\phi_{d}$ are applied to the penultimate-layer features from a ResNet50 model pretrained on ImageNet-1K.
$\phi_{e}$ maps the inputs $\boldsymbol{z}$ into a latent space and $\phi_d$ performs the reconstruction $\phi_d(\phi_e(\boldsymbol{z}))$.
$\phi_{e}$ and $\phi_d$ are optimized by minimizing the differences between inputs and the reconstructions.
To be specific, we choose to minimize the Mean Squared Error (MSE) between $\boldsymbol{z}$ and $\phi_d(\phi_e(\boldsymbol{z}))$ as follows:
\begin{equation}
\label{response:eq:mse}
\min_{\theta_{\phi_e},\theta_{\phi_d}}\sum_{i}\|\boldsymbol{z}_i-\phi_d(\phi_e(\boldsymbol{z}_i))\|_2^2.
\end{equation}
Both the encoder $\phi_e$ and the decoder $\phi_d$ are two-layer multilayer perceptrons (MLPs).
During inference, the negative MSE is adopted as the OoD detection score, as InD inputs are expected to be well reconstructed with low error (MSE) under this optimization setup.
The results are presented in Table \ref{response:tab:kernel-learning} (shallow AE).

\noindent{\bf RFF-OPT} \cite{sinha2016learning}. 
RFF-OPT is specifically optimized based on Random Fourier Features (RFFs) \cite{rahimi2007random}. 
In our KPCA-based framework, RFF is a data-independent kernel approximation technique that plays key roles especially with large-scale data.
In RFF-OPT, an optimal subset of the naive RFFs is learned by solving the following kernel alignment problem given training data $(\boldsymbol{z}_i,y_i)\in\mathbb{R}^m\times\{-1,1\}$:
\begin{equation}
\label{response:eq:alignment}
\max_{\boldsymbol{q}\in{\cal P}_{M_r}}\sum_{i,j}y^iy^j\sum_{n=1}^{M_r}\boldsymbol{q}_n\Phi_{\rm RFF}^n(\boldsymbol{z}_i)\cdot\Phi_{\rm RFF}^n(\boldsymbol{z}_j),
\end{equation}
where ${\cal P}_{M_r}$ is defined as a set of discrete distributions approximating the uniform distribution under the $f$-Divergence ($D_f$) within a constant $\rho$: ${\cal P}_{M_r}\coloneqq\{\boldsymbol{q}:D_f(\boldsymbol{q}||\frac{\bf 1}{M_r}\leq\rho)\}$, and $\Phi_{\rm RFF}^n(\boldsymbol{z}_i)$ denotes the $n$-th element in the mapped RFFs: $\Phi_{\rm RFF}^n(\boldsymbol{z}_i)=\cos({\boldsymbol{\omega}}_n^\top\phi_{\rm cos}(\boldsymbol{z}_i)+u_n)$.
Therefore, Problem (\ref{response:eq:alignment}) aims at finding weights $\boldsymbol{q}$ so that the RFFs-approximated kernel matrix matches the correlation matrix $\boldsymbol{y}\boldsymbol{y}^\top$.
For more details, please kindly refer to \cite{sinha2016learning}.

For a fair comparison, experiments with RFF-OPT are implemented on the penultimate-layer features from a ResNet50 model pretrained on ImageNet-1K, as done in our main comparisons with  RFF-based method in the manuscript.
We follow the public code\footnote{https://github.com/duchi-lab/learning-kernels} of RFF-OPT \cite{sinha2016learning} and extends the bi-classification in Problem (\ref{response:eq:alignment}) to multi-classification for ImageNet-1K.
Due to memory limitation, 10\% of the ImageNet-1K training features are randomly selected and adopted to solve Problem (\ref{response:eq:alignment}).
A sparse subset of the naive RFFs is obtained and applied to our KPCA framework for OoD detection, achieving superior detection performance with reduced computations (fewer mapped dimensions), as demonstrated in Table \ref{response:tab:kernel-learning} (RFF-OPT).

\noindent{\bf DKGP} \cite{wilson2016deep}. Gaussian Processes (GPs) \cite{williams2006gaussian} have been widely utilized and is closely related to kernel methods. In \cite{wilson2016deep}, Deep Kernel GPs (DKGPs) are formulated by applying GPs on the deep-layer features from neural networks, thereby encapsulating the expressive power of deep networks within kernel-based architectures with learnable representations. 
Therefore, similarly in our KPCA framework for OoD detection, the parameters of the Gaussian kernel can be learned through DKGP by maximizing the following Evidence Lower BOund (ELBO) ${\cal L}_{\rm ELBO}$:
\begin{equation}
\label{response:eq:dkgp}
\max_{\gamma_k,\gamma_v}{\cal L}_{\rm ELBO}=\sum_i\mathbb{E}_q\left[\log p(y_i|{\bf g}_i)\right]-{\rm KL}\left[q\|p\right],
\end{equation}
where the optimization parameters include the  kernel parameters $\gamma_k$ and the variational parameters $\gamma_v$. The first term of ${\cal L}_{\rm ELBO}$ is the expected log-likelihood for every sample given the latent function $\bf g$ and the variational distribution $q$.
The second term (KL divergence)  regularizes the approximation to ensure $q$ remaining close to the GP prior $p$.

We introduce DKGP into our KPCA framework to learn the Gaussian kernel.
In experiments, the Gaussian kernel parameters are obtained by solving Problem (\ref{response:eq:dkgp}) on the $\ell_2$-normalized penultimate-layer features from ResNet50 pretrained on ImageNet-1K.
Then, RFF and Nystr\"om mappings based on the learned Gaussian kernel are both implemented to perform KPCA for OoD detection, with results in Table \ref{response:tab:kernel-learning} (RFF-DKGP and Nystr\"om-DKGP).

\noindent{\bf Results and discussions.} Table \ref{response:tab:kernel-learning} compares our proposed KPCA-based framework without (w/o) and with (w/) learnable representations for OoD detection.
The two kernel learning methods, RFF-OPT \cite{sinha2016learning} and DKGP \cite{wilson2016deep}, improve the detection performance.
Both methods optimize our proposed Cosine-Gaussian kernel towards different objectives: the kernel alignment of Problem (\ref{response:eq:alignment}) in RFF-OPT and the classification ELBO of Problem (\ref{response:eq:dkgp}) in DKGP.
Their enhanced detection results with representations learned under varied objectives further validate the flexibility of the Cosine-Gaussian kernel and its effectiveness in capturing the non-linear patterns that separate OoD from InD data.

In contrast, the shallow AE \cite{meissen2022unsupervised} yields weaker detection results. 
This may stem from its optimization objective: the MSE-based objective in Problem (\ref{response:eq:mse}) primarily focuses on feature-level reconstruction fidelity on InD data, and does not explicitly encourage InD-OoD separability.
Unlike our Cosine-Gaussian kernel, which is designed for this separability, the AE learns a latent space optimized for reconstruction rather than discrimination. 
Nevertheless, as an effective learnable non-linear mapping method, AE offers a promising direction for OoD detection through objectives explicitly designed to capture distribution shifts to yield discriminative latent spaces.

\begin{table}[h]
    \centering
    \caption{Detection performance of KPCA with learnable representations. The average FPR and AUROC results over the selected 4 OoD datasets are reported.}
    {\begin{tabular}{@{}l|c|cc@{}}
    \toprule
    \multirow{2}{*}{Approximation} & \multirow{2}{*}{mapped dimension} & \multicolumn{2}{c}{{\bf AVERAGE}} \\
    & & FPR$\downarrow$ & AUROC$\uparrow$ \\
    \midrule
    \multicolumn{4}{c}{\it KPCA w/o learnable representations}\\
    RFF & $M_r$=4,096 & 48.60 & 87.53 \\
    Nystr\"om & $M_n$=2,048 & 38.15 & 91.02 \\
     \midrule
    \multicolumn{4}{c}{\it KPCA w/ learnable representations}\\
    RFF-OPT & $M_r$=1,961 & 46.59 & 87.85 \\
    RFF-DKGP & $M_r$=2,048 & 42.36 & 89.84\\
    Nystr\"om-DKGP & $M_n$=2,048 & 32.34 & 92.51\\
    shallow AE & 2,048 & 60.18 & 82.79 \\
    \bottomrule
    \end{tabular}
    }
    \label{response:tab:kernel-learning}
\end{table}

\end{appendices}

\end{document}